\documentclass{article} %
\usepackage{iclr2026_conference,times}

\usepackage{hyperref}
\usepackage{url}
\usepackage{graphicx}
\usepackage{amsmath,amsfonts,bm}
\usepackage{booktabs}           
\usepackage{multirow}          
\usepackage{subcaption}        
\usepackage[subrefformat=parens]{subcaption} 
\usepackage{mathtools}
\usepackage{enumitem}
\usepackage{xspace}            
\usepackage{calc}              
\usepackage{lineno}            
\usepackage{xcolor}            
\usepackage{microtype} 
\usepackage{tabularx}
\newcommand{\pname}{TDBench\xspace}

\usepackage{xcolor}
\usepackage{soul} %
\soulregister\st{1} %

\newcommand{\best}[1]{\textcolor{green!50!black}{#1}}
\newcommand{\worst}[1]{\textcolor{red!70!black}{#1}}

\newif\ifedit
\edittrue
\definecolor{addcolor}{RGB}{0,153,0}       %
\definecolor{deletecolor}{RGB}{200,0,0}    %

\ifedit
    
    \newcommand{\delete}[2]{%
        {\textcolor{deletecolor}{\textit{#1:} \st{#2}}}%
    }
\else
    
    \newcommand{\delete}[2]{}
\fi

\def\eqref#1{equation~\ref{#1}}

\def\1{\bm{1}}

\DeclareMathAlphabet{\mathsfit}{\encodingdefault}{\sfdefault}{m}{sl}
\SetMathAlphabet{\mathsfit}{bold}{\encodingdefault}{\sfdefault}{bx}{n}

\newcommand{\E}{\mathbb{E}}

\title{TDBench: A Benchmark for Top-Down Image Understanding with Reliability Analysis of Vision-Language Models}

\author{Kaiyuan Hou$^*$, Minghui Zhao$^*$, Lilin Xu, Yuang Fan \& Xiaofan Jiang \\
Department of Electrical Engineering\\
Columbia University\\
\texttt{\{kh3119,mz2869,lx2331,yf2676\}@columbia.edu, jiang@ee.columbia.edu}
}

\iclrfinalcopy %
\begin{document}

\maketitle
\def\thefootnote{$^*$}\footnotetext{These authors contributed equally to this work}\def\thefootnote{\arabic{footnote}}
\begin{abstract}
Top-down images play an important role in safety-critical settings such as autonomous navigation and aerial surveillance, where they provide holistic spatial information that front-view images cannot capture. Despite this, Vision Language Models (VLMs) are mostly trained and evaluated on front-view benchmarks, leaving their performance in the top-down setting poorly understood. Existing evaluations also overlook a unique property of top-down images: their physical meaning is preserved under rotation. In addition, conventional accuracy metrics can be misleading, since they are often inflated by hallucinations or ``lucky guesses'', which obscures a model’s true reliability and its grounding in visual evidence. To address these issues, we introduce TDBench, a benchmark for top-down image understanding that includes 2000 curated questions for each rotation. We further propose RotationalEval (RE), which measures whether models provide consistent answers across four rotated views of the same scene, and we develop a reliability framework that separates genuine knowledge from chance. Finally, we conduct four case studies targeting underexplored real-world challenges. By combining rigorous evaluation with reliability metrics, TDBench not only benchmarks VLMs in top-down perception but also provides a new perspective on trustworthiness, guiding the development of more robust and grounded AI systems. \footnote{Project website: https://github.com/Columbia-ICSL/TDBench}
\end{abstract}

\section{Introduction}\label{sec:intro}
Top-down images provide comprehensive spatial overviews and clear geometric context, supporting tasks such as autonomous navigation, aerial surveillance, mapping, and disaster assessment~\citep{Lu2018UAV, Nearmap2022aerial, zhao2025flexifly}. Top-down images from drones or satellites provide a complete ``bird's-eye'' view, offering several unique advantages over conventional front-view images: they reduce occlusion between objects, maintain more consistent scale across the frame, and reveal complete spatial layouts that are impossible to observe from ground level. These properties allow analysts or autonomous systems to reason about large geographic areas efficiently, which is essential in applications such as traffic monitoring, urban planning, and environmental response.

Despite their importance, top-down images are substantially underrepresented in the datasets commonly used to train and evaluate Vision Language Models (VLMs). Well-known datasets such as COCO~\citep{lin2015coco} and ImageNet~\citep{russakovsky2015imagenet} contain primarily front-view images, where appearance cues, object sizes, and spatial relationships are largely different from aerial perspectives. For instance, in our preliminary data audit, fewer than 7\% images (595 of 8,629) from the VisDrone dataset~\citep{zhu2021visdrone} could be considered truly top-down. This limited coverage leaves current VLMs largely untested for top-down understanding, even though such models are increasingly applied in drone-based and remote-sensing systems.

Most existing VLM benchmarks~\citep{liu2024mmbench,yue2024mmmu,yu2024mmvet,lu2024mathvista}
are not designed for top-down images. While these benchmarks have driven progress in general-purpose visual reasoning, they provide little insight into how VLMs handle the distinct challenges of top-down perception. Aerial scenes present small, densely packed objects, drastically different viewing angles, and weak perspective depth cues. Contextual cues that aid object recognition in conventional images may be absent or transformed in top-down perspectives. VLMs trained mostly on canonical-view data often fail to generalize to these conditions, leading to severe accuracy drops~\citep{danish2025geobench, li2024topviewrs}. Without a dedicated benchmark, it is difficult to measure or systematically improve their performance on top-down views.

To address this gap, we present \pname, a benchmark for evaluating VLMs on top-down image understanding. \pname contains 2,000 carefully constructed questions drawn from public aerial datasets and high-fidelity simulations, covering diverse settings and tasks relevant to real-world operations. We also introduce RotationalEval (RE), an evaluation method that leverages a key property of top-down images: their physical meaning is preserved under rotation. Unlike front-view images, where rotation produces implausible scenes (for example, the sky appearing below or objects upside down), rotating a top-down image is equivalent to changing a drone’s heading, so the scene remains physically consistent. RE tests whether models can answer correctly across all four rotated views, recognizing that semantics and object identities remain the same while spatial descriptors (e.g., ``top-left''), and coordinates legitimately change. This provides a stricter and more diagnostic measure of visual reasoning, reducing the influence of spurious one-off successes.

Vision Language Models (VLMs) often hallucinate, generating answers from learned text patterns instead of grounding them in the provided image~\citep{li2023hallucination, bai2025hallucination}. This can artificially inflate scores under conventional evaluation. However, an ungrounded guess is highly unlikely to be correct across four different rotations, RE naturally filters out these successes. We further formalize this with new reliability-oriented metrics that disentangle a model's visually-grounded knowledge from its apparent accuracy. This provides a more quantitative view of model trustworthiness than raw accuracy alone.
\begin{figure}[t]
  \centering
  \captionsetup{skip=3pt}
  \includegraphics[width=0.5\linewidth]{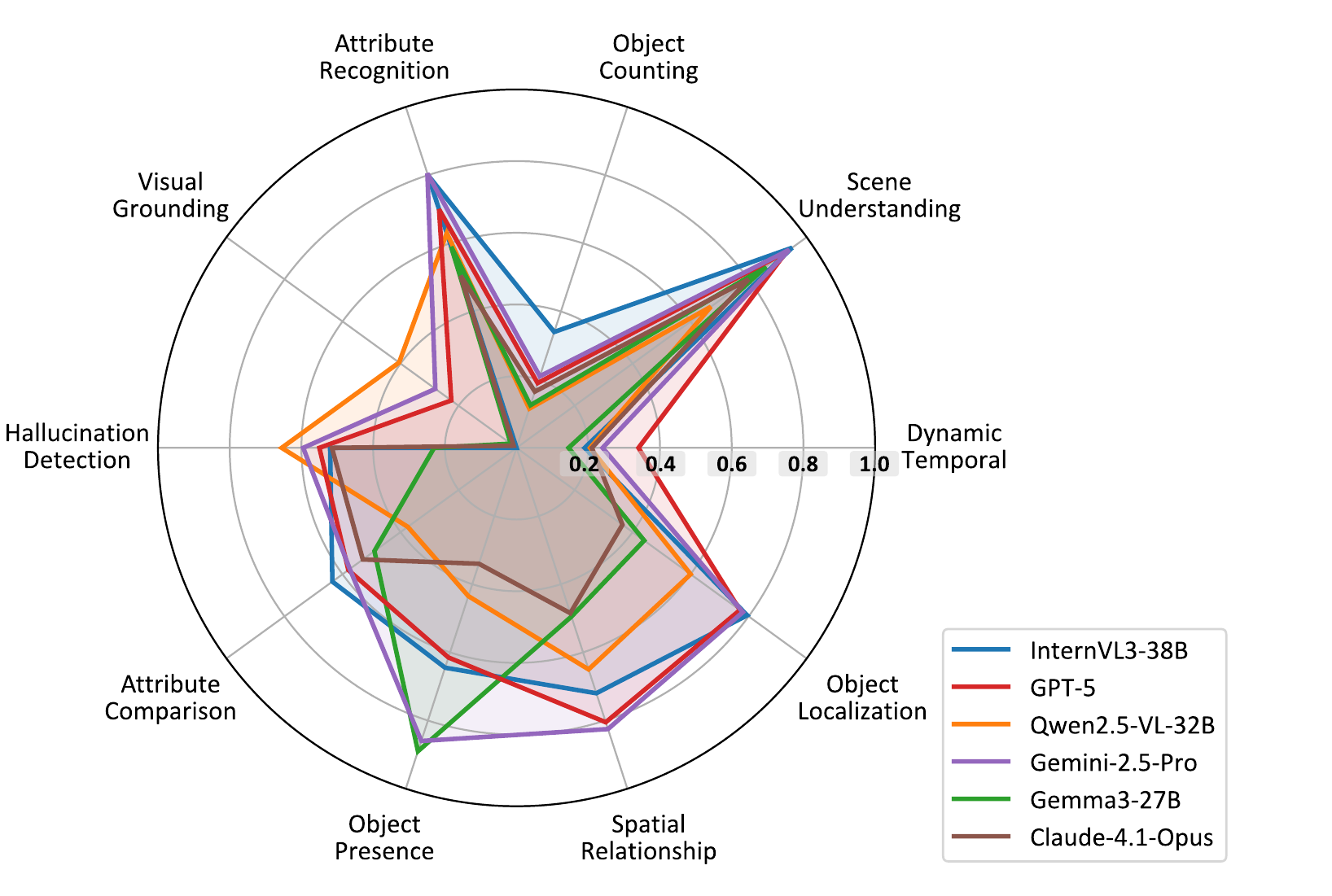}
  \hfill
  \includegraphics[width=0.48\linewidth]{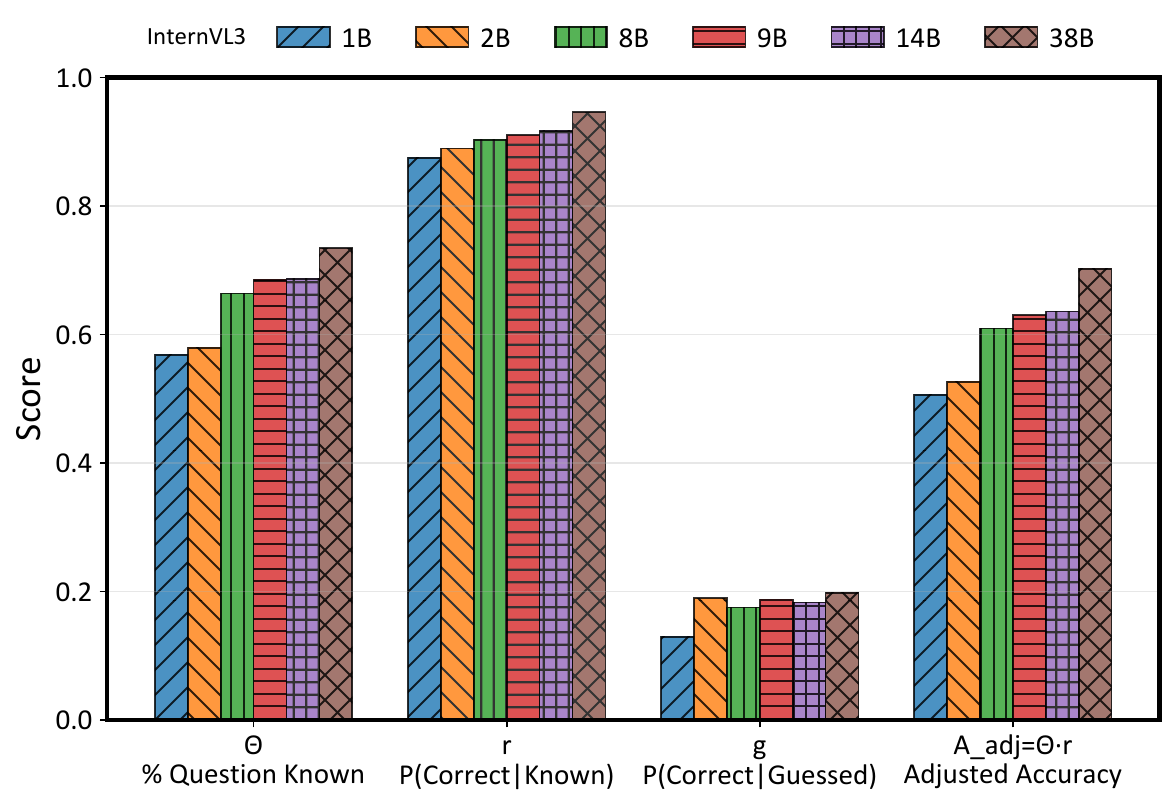}
  \caption{\textbf{(Left)} Accuracy across ten top-down image tasks in \pname. \textbf{(Right)} Knowledge decomposition analysis from \pname: \textit{\% of questions known ($\theta$)} measures the proportion of questions a model truly knows; \textit{$P(Correct\vert Known)$ ($r$)} is the model's accuracy among the questions that it knows; \textit{$P(Correct\vert Guessed)$ ($g$)} is the model's accuracy among the questions it does not know; and the \textit{Adjusted Accuracy ($A_{\text{adj}}=\theta\cdot r$)} is the model's accuracy without lucky guesses.}
  \label{fig:overall_performance}
  \vspace{-0.2in}
\end{figure}

Finally, we conduct four application-oriented case studies for real-world applications: digital and physical ``zoom-in'', handling partially visible objects and reasoning about depth from 2D views. These case studies demonstrate how \pname can guide the design and deployment of VLM-based aerial systems. In summary, our main contributions are:

\begin{itemize}[leftmargin=*]
\item \textbf{Application-driven Benchmark.} We build \pname, a top-down benchmark of \textbf{2{,}000} question–answer pairs from public datasets and high-fidelity simulation, organized into ten evaluation dimensions. To demonstrate its practical relevance, we also conduct \textbf{four case studies} that examine VLMs on real-world aerial applications, providing actionable insights for deployment.

\item \textbf{Rotation-invariant Evaluation.} We introduce \textbf{RotationalEval (RE)}, an evaluation strategy that requires consistent answers across four rotated views of each image. By requiring models to be rotationally consistent, correctly adapting their spatial reasoning to each orientation, RE provides a far more robust and diagnostic measure of their performance than single-view evaluation.

\item \textbf{Probability-based Knowledge Reliability Analysis.} Beyond raw and RE accuracy, we propose a \textbf{probabilistic analysis} that decomposes model performance into
\textit{\% of questions known ($\theta$)}, \textit{$P(Correct\vert Known)$ ($r$)}, \textit{$P(Correct\vert Guessed)$ ($g$)}, and further aggregate them into \textit{Adjusted Accuracy ($\theta\cdot r$)}, which reveals how much of a model’s apparent correctness stems from genuine knowledge rather than lucky guesses.
\end{itemize}

\section{Related Works}\label{sec:related}
\subsection{Vision Language Models (VLMs)}
Vision Language Models (VLMs) extend large language models (LLMs) to visual inputs by aligning image features with text representations. Most current VLMs adopt a two-stage design: a pretrained visual encoder (e.g., CLIP~\citep{radford2021clip} or SigLIP~\citep{zhai2023siglip}) is coupled with a pretrained text-only LLM via a learnable projection module, as in LLaVA~\citep{li2024llava} and InternVL~\citep{chen2025internvl25}. This setup preserves the language backbone while enabling it to interpret visual features. Some models instead use early-fusion architectures that train perception and language components jointly, strengthening visual grounding and cross-modal reasoning. Proprietary models such as GPT~\citep{openai2024gpt4}, Gemini~\citep{geminiteam2024gemini}, and Claude~\citep{anthropic2024claude35sonnet} may follow similar multimodal principles at larger scales.

VLMs are generally trained on large-scale image–text pairs from datasets like LAION~\citep{schuhmann2022laion5b}, COCO~\citep{lin2015coco}, and ImageNet~\citep{russakovsky2015imagenet}, which may contain few top-down images and thus treat them as out-of-distribution (OOD). While this broad training enables rich visual–linguistic knowledge, it biases models toward ground-level scenes and object appearances. As a result, their generalization to top-down views, where objects appear smaller, depth cues are weak, and spatial relationships dominate, remains underexplored, motivating the need for a dedicated benchmark.

\subsection{VLM Benchmarks}
Recent years have seen the emergence of numerous benchmarks for evaluating Vision–Language Models (VLMs) on diverse multimodal reasoning tasks. General-purpose benchmarks such as MMBench~\citep{liu2024mmbench}, MMMU~\citep{yue2024mmmu}, MME~\citep{fu2024mme}, and MM-Vet~\citep{yu2024mmvet} assess general knowledge, visual perception, commonsense reasoning, and spatial understanding. However, these benchmarks focus primarily on conventional front-view imagery and include few tasks involving aerial or top-down perspectives. They thus overlook challenges unique to top-down understanding, including extreme scale variation, weak depth cues, and dense spatial layouts, which often cause VLMs to underperform on aerial tasks.

A few recent efforts have begun addressing this gap using remote sensing images. For example, \citet{hu2023rsgpt}, \citet{muhtar2024lhrs}, \citet{kuckreja2023geochat}, and \citet{danish2025geobench} evaluate VLMs on satellite data. These datasets mostly comprise low-resolution images (meters per pixel) aimed at large-scale land cover classification or scene categorization. They rarely involve human-scale and near-surface views tasks such as object localization, attribute comparison, or spatio-temporal analysis. Moreover, satellite images are typically captured from fixed nadir viewpoints at consistent altitudes, lacking the perspective variation and dynamic conditions common in drone operations.

Beyond remote sensing, only a few studies explore top-down images. For instance, \citet{li2024topviewrs} introduces an indoor map benchmark for evaluating navigation and spatial reasoning from floor plans. In contrast, our benchmark \pname focuses on high-resolution, near-surface top-down images resembling drone viewpoints, enabling systematic evaluation of fine-grained perception and reasoning abilities that remain underrepresented in existing benchmarks.

\subsection{Hallucinations in Multimodal LLMs}
Hallucination has become an increasing concern in both large language models (LLMs) and vision–language models (VLMs). In VLMs, it often occurs when models generate content that is inconsistent with the image, such as describing nonexistent objects, misrepresenting spatial relationships, or ignoring the visual input entirely~\citep{wang2024mllmseedynamiccorrection}. Recent studies have introduced benchmarks and methods to systematically evaluate these visual hallucinations. \citet{li2023evaluating} introduced the POPE method, which probes object hallucination by asking targeted presence/absence questions and measuring how often models falsely claim the existence of unseen objects. \citet{liu2024hallucination} provided a large-scale study on hallucinations in VLMs and proposed automatic detection metrics based on grounding scores, which assess alignment between textual output and visual evidence. HallusionBench~\citep{guan2024hallusionbench} proposed a diagnostic benchmark designed to isolate hallucination behavior using paired, contrastive visual questions to reveal when models invent objects or attributes. 

These approaches typically rely on comparing generated captions or answers against ground-truth annotations, using measures such as hallucination rate (percentage of fabricated objects), grounding accuracy (percentage of correctly grounded mentions), or contrastive consistency scores. However, current methods primarily treat hallucination as a binary outcome (hallucinated or correct) and do not assess whether correct answers arise from genuine visual understanding or from chance agreement with priors. Our benchmark \pname complements these efforts by a reliability-oriented evaluation perspective, aiming to distinguish reliably grounded responses from lucky successes.

\section{Design of \pname}\label{sec:design}
In this section, we provide an overview of the 10 categories in \pname with examples in Figure~\ref{fig:benchmark_example}. We then describe the data sources used to build the benchmark and the procedures for curating and annotating the dataset.

\subsection{Ability Taxonomy of \pname}
\begin{figure}[t]
\begin{center}
    \includegraphics[width=\linewidth]{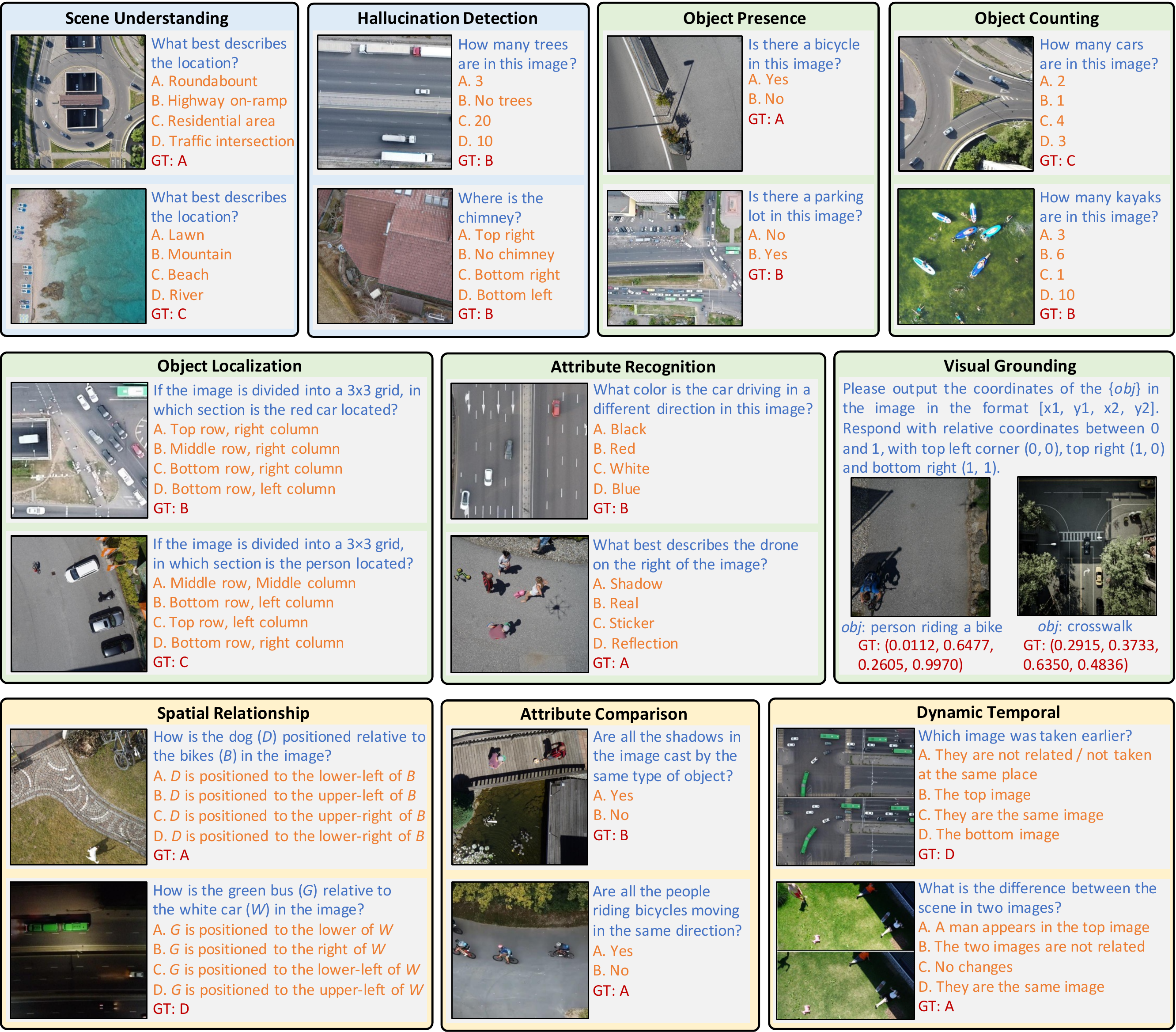}
\end{center}
\caption{Benchmark examples across the ten categories in \pname. Different colors indicate the three high-level capability groups: image perception (blue), single-instance understanding (green), and multi-instance reasoning (yellow). `GT' refers to ground truth.}
\label{fig:benchmark_example}  
\end{figure}
\pname evaluates top-down image understanding across 10 categories, as illustrated in Figure~\ref{fig:overall_performance} (Left). These categories are derived from common aerial tasks in real-world applications. To ensure the benchmark targets perspective-sensitive capabilities, we excluded evaluation dimensions that are already well covered in existing benchmarks or that remain largely unaffected by viewpoint changes, such as text recognition or general knowledge recall. Broadly, the categories can be grouped into three types:

\paragraph{Image Perception}
This category focuses on the broad-scale interpretation of top-down aerial imagery, emphasizing holistic semantic understanding rather than fine-grained details. Such capabilities are especially valuable for wide-area reconnaissance, where drones must scan large regions to detect critical features such as wildfire outbreaks, traffic congestion, or emergency response scenarios. It includes two tasks: \textit{Scene Understanding}, which evaluates a model’s ability to comprehend the overall contextual meaning of a scene, and \textit{Hallucination Detection}, which assesses its ability to distinguish actual image content from fabricated choices. These tasks are shown in \textbf{blue} in Figure~\ref{fig:benchmark_example} and represent foundational abilities for reliable aerial image interpretation.

\paragraph{Single-Instance Understanding}
This category emphasizes detailed object-level recognition and localization within a single image, as shown in \textbf{green} in Figure~\ref{fig:benchmark_example}. It covers both recognition and localization aspects. For recognition, \textit{Object Presence} evaluates basic detection capabilities, and \textit{Attribute Recognition} assesses the identification of specific properties such as color, shape, material, or species. For localization, we use a three-tiered approach: coarse presence detection (\textit{Object Presence}), intermediate 3×3 grid-based localization (\textit{Object Localization}) requiring quadrant-level precision, and fine-grained \textit{Visual Grounding} using exact bounding box coordinates. We also include \textit{Object Counting} to assess quantification abilities, which is particularly challenging in aerial contexts where many similar objects appear at varying scales and densities.

\paragraph{Multi-Instance Reasoning}
This category evaluates compositional reasoning across multiple objects, requiring analysis of spatial, comparative, and temporal relationships, as shown in \textbf{yellow} in Figure~\ref{fig:benchmark_example}. \textit{Spatial Relationship} tasks challenge models to localize multiple objects and accurately determine their relative positions, which is crucial for navigation and path planning in autonomous aerial systems. \textit{Attribute Comparison} requires models to compare properties or states across multiple entities, useful for anomaly detection and identifying distinctive features. Finally, \textit{Dynamic Temporal} presents pairs of images to evaluate models’ ability to detect changes, reason about temporal order, and infer causal relationships.

\subsection{\pname Construction}
\subsubsection{Data Sources}
We constructed \pname from two primary sources: curated public datasets~\citep{roboflow2025shaha, zhu2021visdrone, gasienica2021ensemble, tugraz2025drone, varga2022seadronessee, eradataset} and realistic simulation (CARLA Simulator~\citep{Dosovitskiy17} and GTA V ) covering varied environments, including urban infrastructure, remote wilderness, and disaster zones (Table~\ref{tab:data_distribution}). All images from these datasets were manually selected and annotated following our evaluation taxonomy. In addition to real-world data, we generated synthetic images using the CARLA simulator with custom scripts to control scene parameters precisely. For specialized case studies requiring exact ground truth, such as camera altitude, object counts, or height measurements, we used both CARLA and \textit{Grand Theft Auto V (GTA V)}.

\begin{table}[t]
\centering
\caption{Distribution of data sources in \pname}
\label{tab:data_distribution}
\begin{tabular}{cccc}
\hline
Image Source & Problem Formulation & \textbf{Number} & Ratio \\
\hline
Aerial Traffic Images~\citep{roboflow2025shaha} & Human Annotation & \textbf{457} & 20.8\% \\
Semantic Drone~\citep{tugraz2025drone} & Human Annotation & \textbf{653} & 29.7\% \\
AFO~\citep{gasienica2021ensemble} & Human Annotation & \textbf{18} & 0.8\% \\
Visdrone~\citep{zhu2021visdrone} & Human Annotation & \textbf{416} & 18.9\% \\
Seadronesee~\citep{varga2022seadronessee} & Human Annotation & \textbf{3} & 0.1\% \\
ERA~\citep{eradataset} & Human Annotation & \textbf{363} & 16.5\% \\
CARLA~\citep{Dosovitskiy17} & Simulation Script & \textbf{290} & 13.2\% \\
\hline
\multicolumn{4}{c}{\textbf{Additional New Data Used In Case Study}} \\
\hline
CARLA~\citep{Dosovitskiy17} & Simulation Script & \textbf{1500} & - \\
GTA V & Human Annotation & \textbf{400} & - \\
\hline
\end{tabular}
\end{table}

\paragraph{CARLA Simulation}
CARLA~\citep{Dosovitskiy17} is an open-source autonomous driving simulator that provides high-fidelity urban environments and physics. We used its configurable RGB and segmentation cameras at various altitudes to generate synthetic data. This setup enables precise control over object instances (e.g., vehicles), supporting systematic evaluation of object counting performance (Section~\ref{sec:design}) and altitude-dependent detection studies (Section~\ref{sec:case_study}).

\subsubsection{Data Implementation}
\pname includes two task types: Multiple Choice Questions (MCQs) for most abilities, and Visual Grounding (VG). Each MCQ problem is structured as a quadruple $P_i = [Q_i, I_i, C_i, L_i]$, where $Q_i$ denotes the textual question, $I_i$ is the associated image, $C_i$ represents the set of possible answers with $n$ ($2 \leq n \leq 4$) choices $\{c_1, c_2, \ldots, c_n\}$ (randomly shuffled during evaluation), and $L_i$ is the correct label. For VG problems, we evaluate models' ability to precisely localize objects by comparing their predicted bounding box coordinates against $L_i$, which contains human-annotated ground truth coordinates.

\paragraph{Image Standardization}
To mitigate evaluation biases from inconsistent image preprocessing across different VLMs (such as padding, stretching, or multi-tiling), we established a uniform input pipeline. All images were standardized to a fixed 512×512 pixel resolution. For tasks requiring image pairs, such as temporal or comparative analyses, we concatenated two sub-images either horizontally (as a 512×256 pair) or vertically (as a 256×512 pair). This method ensures the combined input fits the same 512×512 canvas, providing a fair and consistent basis for model comparison.

\paragraph{Quality Control}
We followed a two-stage quality control pipeline combining human and model-based checks.  
\textit{Stage 1: Human review.} Six annotators independently examined all questions, removing or revising questions that were unsolvable due to lost context in the question, or that contained unclear wording or incorrect ground truth.  
\textit{Stage 2: Model filtering.} Several open-source models (Deepseek VL2-small, InternVL2.5-MPO, Phi4, and Qwen2.5-VL 7B) were benchmarked to identify questions that were consistently failed or consistently solved across all models. Questions that all models failed underwent additional human review and were retained only if they were correctly formulated. Conversely, questions that all models solved were removed from \pname since they offered little discriminative value for model comparison.

\subsection{\pname Statistics}
\pname contains 2000 problems across the 10 ability categories for each rotation, plus an additional 2100 problems used in four case studies. We aimed for an even distribution of problems across abilities, with 200 samples per category. Of the total questions, 1910 (including case studies) are collected from real-world datasets, and 2190 are generated from simulation environments. Notably, all problems in the `Object Counting' category are generated from the CARLA Simulator, which allows controlled ground-truth labeling during scene generation. Under RotationalEval (RE), each question is evaluated across four orientations, effectively producing four instances per problem.

\subsection{Leveraging Rotational Invariance in Evaluation}
\begin{figure}[t]
\centering
\captionsetup{skip=3pt}
\includegraphics[width=1\linewidth]{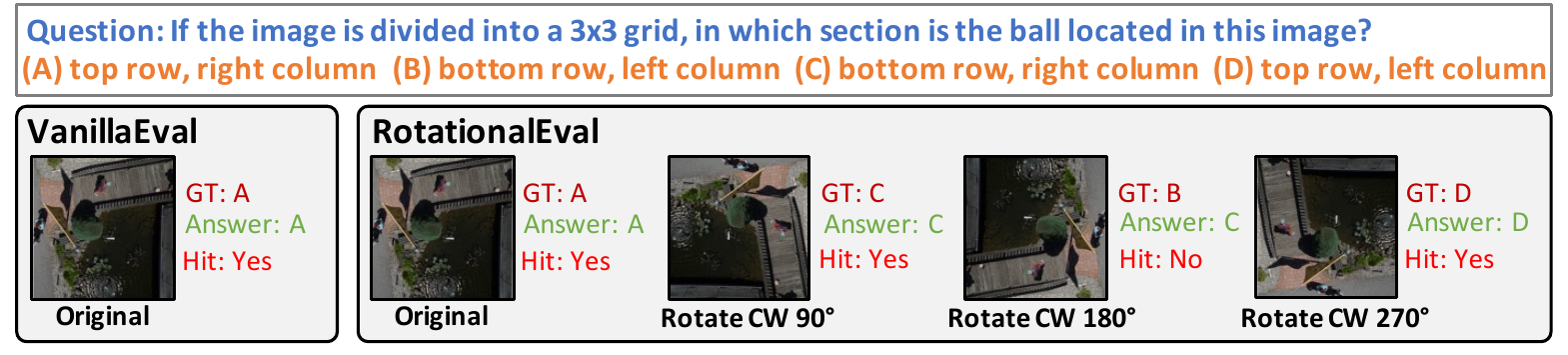}
\caption{\textbf{Proposed RotationalEval (RE) strategy.} In RE, each image is rotated three times to create four questions, with choices generated separately for each rotation. We illustrate a failure case in \textit{object localization} where four choices align with four images, and the VLM answers three correctly but fails on one. }
\label{fig:rot_eval}  
\end{figure}

In \pname, we introduce a novel evaluation strategy, \textbf{RotationalEval (RE)}, designed to leverage the unique properties of top-down images (Figure~\ref{fig:rot_eval}, example from \textit{object localization}). RE evaluates model performance on four orientations of each image: the original, $90^\circ$, $180^\circ$, and $270^\circ$ rotations, and counts a question as correct only if \textbf{all} four are answered correctly. This exploits the fundamental \textbf{rotational invariance} of aerial perspectives: unlike front-view images, where orientation conveys semantic information, top-down images preserve their meaning across rotations. Such rotations simply mimic different yaw angles during capture without altering scene content or physical spatial relationships. This property enables more rigorous evaluation, which would be unsuitable for front-view images where rotations create physically implausible scenes. 

\paragraph{Rotation-Aware Question Design}
Because \pname supports RotationalEval (RE), we categorized all questions as either \textbf{rotation-invariant} or \textbf{rotation-sensitive}.  
Rotation-invariant questions (e.g., object presence, attribute recognition) remain semantically unchanged after rotation; only the image is rotated while the question and answer options remain the same.  
Rotation-sensitive questions (e.g., spatial relationships or localization) require synchronized transformation of directional references. For instance, after a 90$^\circ$ clockwise rotation, phrases like ``top-left'' are mapped to ``top-right''.  To automate this process, we use placeholder tokens (\(\langle\)img1\(\rangle\), \(\langle\)img2\(\rangle\)) in both questions and answers. In the original orientation, they are rendered as ``left/right'' or ``top/bottom'', and these tokens are automatically rotated when generating the 90$^\circ$, 180$^\circ$, and 270$^\circ$ variants. This ensures consistent semantics across all rotation conditions.

\section{Evaluation Results}\label{sec:results}

\subsection{Setup}
To ensure reproducibility and a fair comparison across models, all evaluations are conducted with VLMEvalKit~\citep{duan2024vlmevalkit}, an open-source VLM evaluation framework. We evaluated a total of 60 VLMs in a zero-shot setting, without providing any in-context examples. For all experiments, the model temperature was set to 0, and GPT-4o was used as the answer extractor for all model outputs.

\paragraph{Models}We evaluated 17 \textbf{proprietary models}, including the Claude~\citep{anthropic2024claude35sonnet, anthropic2024claude37sonnet, anthropic2024claude4sonnet, anthropic2024claude41opus}, Gemini~\citep{geminiteam2024gemini, geminiteam2025gemini}, and GPT~\citep{openai2024gpt4, openai2025gpt41, openai2025gpto3, openai2025gpt5} families; and 43 \textbf{open-source models} from diverse families such as Gemma 3~\citep{gemmateam2025gemma3technicalreport}, InternVL~\citep{chen2025internvl25, zhu2025internvl3, wang2025internvl35}, Qwen2.5-VL~\citep{Qwen2.5-VL}, DeepSeek VL2~\citep{wu2024deepseekvl2}, LLaVA~\citep{liu2023llava,li2024llava}, Kimi-VL~\citep{kimiteam2025kimivltechnicalreport}, and VLM-R1~\citep{shen2025vlmr1stablegeneralizabler1style}. These models span a wide range of sizes, from 0.5 billion to 38 billion parameters.

\begin{table}[b]
\centering
\captionsetup{skip=1pt}
\caption{Performance comparison of open-source and proprietary VLMs under VanillaEval (VE@$0^\circ$) and RotationalEval (RE), along with the corresponding accuracy drop ($\Delta$) on \pname.}
\label{tab:rotationalEval}
\resizebox{1.\linewidth}{!}{
\begin{tabular}{lrrr|lrrr}
\toprule
\textbf{Open VLMs} & \textbf{VE} & \textbf{RE} & \textbf{$\Delta$} & \textbf{Prop VLMs} & \textbf{VE} & \textbf{RE} & \textbf{$\Delta$} \\
\midrule
Qwen2.5-VL 7B      & 0.630 & \textbf{0.470} & \textcolor{red}{-0.160} & Gemini 2.5 Pro     & \textbf{0.793} & \textbf{0.611} & \textcolor{red}{-0.182} \\
Kimi-VL            & 0.624 & 0.455 & \textcolor{red}{-0.169} & Gemini 1.5 Pro     & 0.756 & 0.572 & \textcolor{red}{-0.183} \\
DeepSeek VL2       & \textbf{0.637} & 0.448 & \textcolor{red}{-0.189} & GPT-5              & 0.761 & 0.570 & \textcolor{red}{-0.190} \\
InternVL3.5 14B    & 0.601 & 0.442 & \textcolor{red}{-0.159} & GPT-4.1            & 0.720 & 0.520 & \textcolor{red}{-0.200} \\
LLaVA-Next-13B     & 0.617 & 0.419 & \textcolor{red}{-0.198} & Claude Sonnet 3.7  & 0.611 & 0.415 & \textcolor{red}{-0.196} \\
Gemma3 12B         & 0.591 & 0.330 & \textcolor{red}{-0.260} & Claude Opus 4.1    & 0.603 & 0.392 & \textcolor{red}{-0.211} \\
\bottomrule
\end{tabular}
}
\end{table}

\subsection{Results}
\paragraph{RotationalEval vs. VanillaEval}
We first compare our proposed RotationalEval (RE) with the conventional one-pass evaluation, VanillaEval (VE). In this experiment, we use images without any rotations, denoted as VE@$0^\circ$. Table~\ref{tab:rotationalEval} summarizes their results on \pname, averaged across all dimensions. Adopting RE leads to a notable performance decline across all VLMs. This drop occurs because RE reduces the chance of obtaining correct answers through random guessing. Interestingly, models with higher VE do not necessarily achieve higher RE. For example, although Gemini~1.5~Pro has a slightly lower VE than GPT-5 (0.756 vs.\ 0.761), it attains a higher RE (0.572 vs.\ 0.570). Among open models, DeepSeek~VL2 achieves the best VE, while Qwen2.5-VL-7B achieves the highest RE. These results suggest that models performing well under VE may still be prone to hallucinations, which we further examine in Section~\ref{sec:robustness}.

\paragraph{Main Results}
All reported results are based on \textbf{RotationalEval (RE)}, calculated as the \textit{average} across ten evaluation categories unless explicitly stated. Detailed results, including \textit{dimension-wise performance}, are provided in Appendix~\ref{app:additional_eval}.

Figure~\ref{fig:re_performance_opensource} shows the RE performance of various open-source models as a function of their parameter size. Within the same model families, performance generally increases with model size, although several exceptions exist. Notably, the \textit{``thinking'' variants consistently underperform their standard counterparts}, especially at smaller model sizes, with the gap narrowing as model size increases. This suggests that while chain-of-thought prompting can enhance reasoning at the semantic level, it may make responses less grounded in the visual input. 
In addition, newer models do not necessarily perform better: for example, InternVL3.5 underperforms InternVL3 despite being trained on more data, suggesting that additional general-purpose data may have diluted the proportion of top-down-related images during training. We also report the performance of proprietary models in Figure~\ref{fig:re_performance_proprietary}; although their parameter sizes are undisclosed, the largest variants generally outperform their smaller counterparts, except for GPT-4.1 and GPT-4.1-Nano.

\begin{figure}[t]
    \centering
    \captionsetup{skip=3pt}
    \begin{subfigure}[b]{0.7\textwidth}
        \centering
        \captionsetup{skip=0pt}
        \includegraphics[width=\linewidth]{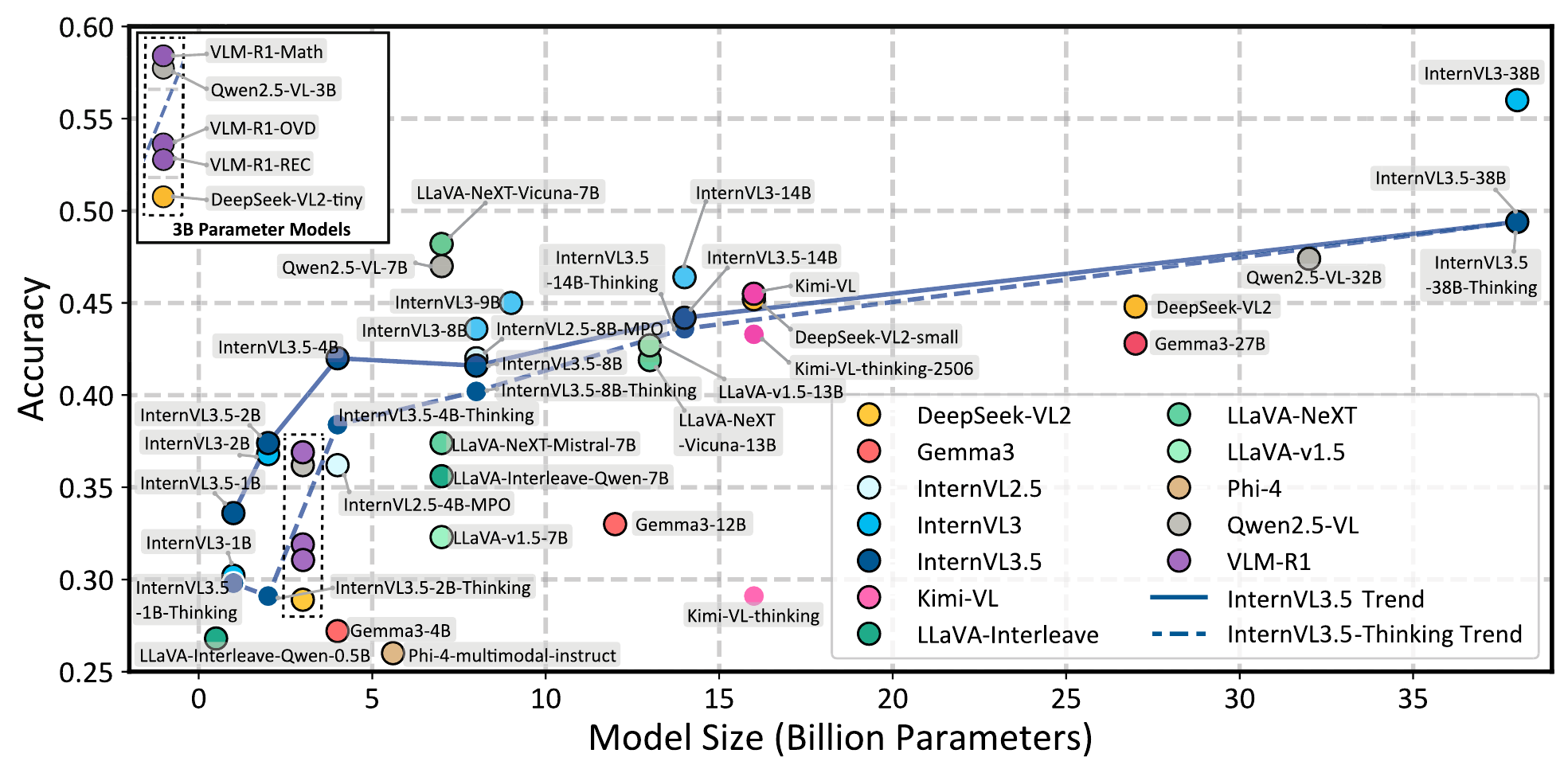}
        \caption{Open-source models.}
        \label{fig:re_performance_opensource}
    \end{subfigure}
    \hfill
    \begin{subfigure}[b]{0.292\textwidth}
        \centering
        \captionsetup{skip=0pt}
         \includegraphics[width=\textwidth]{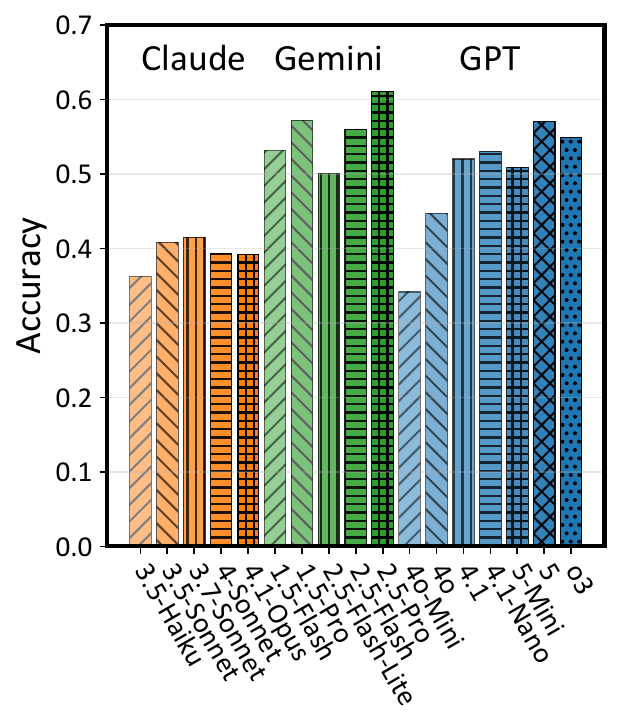}
        \caption{Proprietary models.}
        \label{fig:re_performance_proprietary}
    \end{subfigure}
    \caption{Average RE performance of models on \pname, aggregated across 10 evaluation dimensions for both Open-source and Proprietary models.}
    \label{fig:re_performance}
\end{figure}

\subsection{Beyond Accuracy: A Deeper Analysis of Model Reliability}
\label{sec:robustness}

\begin{table}[b!]
\centering
\captionsetup{skip=1pt}
\caption{RE, MA, $\overline{\text{VE}}$, and reliability parameters (proportion of questions a model truly knows $\theta$, accuracy among known questions $r$, accuracy among guessed questions $g$, and adjusted accuracy $A_{\text{adj}}$). Arrows indicate whether higher ($\uparrow$) or lower ($\downarrow$) is better. Best values are \textcolor{green!50!black}{green}, worst are \textcolor{red!70!black}{red}.}
\label{tab:model_robustness}
\setlength{\tabcolsep}{6pt}
\renewcommand{\arraystretch}{1.15}
\resizebox{0.9\linewidth}{!}{
\begin{tabular}{l|c c c|c c c c}
\hline
\textbf{Model} & \textbf{RE}$\uparrow$ & \textbf{MA}$\downarrow$ & \textbf{$\overline{\text{VE}}$}$\uparrow$ & $\mathbf{\theta}$$\uparrow$ & $\mathbf{r}$$\uparrow$ & $\mathbf{g}$\ & \textbf{$A_{\text{adj}}$}$\uparrow$ \\ \hline
Gemini 2.5 Pro & \textbf{\textcolor{green!50!black}{0.611}} & \textbf{\textcolor{green!50!black}{0.073}} &\textbf{\textcolor{green!50!black}{0.791}} & \textbf{\textcolor{green!50!black}{0.822}} & 0.909 & 0.201 & \textbf{\textcolor{green!50!black}{0.754}} \\
GPT-5 & 0.570 & 0.085 & 0.751 & 0.688 & \textbf{\textcolor{green!50!black}{0.941}} & 0.265 & 0.652 \\
Claude Opus 4.1 & \textbf{\textcolor{red!70!black}{0.392}} & 0.194 & 0.607 & 0.610 & \textbf{\textcolor{red!70!black}{0.849}} & 0.189 & 0.541 \\
o3 & 0.549 & 0.096 & 0.731 & 0.693 & 0.921 & 0.279 & 0.651 \\
DeepSeek VL2 & 0.448 & 0.196 & 0.631 & 0.620 & 0.900 & 0.184 & 0.568 \\
Gemma3-27B & 0.428 & 0.220 & \textbf{\textcolor{red!70!black}{0.604}} & \textbf{\textcolor{red!70!black}{0.587}} & 0.880 & 0.206 & \textbf{\textcolor{red!70!black}{0.538}} \\
Qwen2.5-VL-32B & 0.474 & 0.165 & 0.668 & 0.668 & 0.902 & 0.203 & 0.611 \\
Kimi-VL & 0.455 & \textbf{\textcolor{red!70!black}{0.239}} & 0.613 & 0.612 & 0.882 & 0.164 & 0.565 \\ \hline
\end{tabular}}
\end{table}

As noted earlier, RotationalEval (RE) yields lower scores than 
VanillaEval (VE) 
because it discounts isolated correct predictions and thus reduces the impact of lucky guesses. To further analyze this phenomenon, let $\Phi=\{0^\circ,90^\circ,180^\circ,270^\circ\}$ be the set of rotations, and let $Y_{i}^{(\phi)}\in\{0,1\}$ denote whether the question $i$ under rotation $\phi$ is answered correctly. We define three observations
\[
\mathrm{RE}=\Pr(\forall \phi\in\Phi:Y_{i}^{(\phi)}=1),\quad
\overline{\mathrm{VE}}=\E\Big[\frac{1}{|\Phi|}\sum_{\phi\in\Phi}Y_{i}^{(\phi)}\Big],\quad
\mathrm{MA}=\Pr(\forall \phi:Y_{i}^{(\phi)}=0),
\]
where MA denotes wrong answer in all rotations.
Assuming each question for the model is either ``known'' or ``unknown'', and rotations are conditionally independent, the above observations satisfy
\[
\mathrm{RE}=\theta r^4+(1-\theta)g^4,\qquad
\overline{\mathrm{VE}}=\theta r+(1-\theta)g,\qquad
\mathrm{MA}=\theta(1-r)^4+(1-\theta)(1-g)^4.
\]
where $\theta$ represents \textit{the proportion of questions the model truly knows}, $r$ means \textit{the accuracy on known questions}, and $g$ denotes \textit{the accuracy on unknown questions} (due to lucky guesses). These parameters are inferred by solving the system of equations above (see Appendix~\ref{app:identifiability} for derivation); We aggregate these into the adjusted accuracy ($A_{\text{adj}}$):
\[
\ A_{\text{adj}} = \theta \cdot r.
\]
The adjusted accuracy represents single-pass accuracy after discounting the contribution of guessing from the apparent correctness (VE). To illustrate this, Figure~\ref{fig:overall_performance} (Right) presents results for different sizes of InternVL3, averaged across all evaluation dimensions. As model size increases, $\theta$ (the proportion of questions the model truly knows) also rises, while $r$ remains consistently high (approaching 100\%), and shows a gradual upward trend with scale, which is desirable. In contrast, $g$ exhibits variability that does not show a clear dependence on model size. Overall, Adjusted Accuracy improves with larger models, supporting the validity of our probability-based knowledge reliability analysis. 
Table~\ref{tab:model_robustness} reports additional representative results across different model families (including four proprietary and four open-source VLMs), with full category-wise breakdowns for all 60 models provided in Appendix~\ref{app:additional_eval}.

Unlike the scaling trend observed with InternVL3, different models exhibit distinct strengths and weaknesses on \pname. For example, \textit{Gemini 2.5 Pro} achieves the highest $\theta$, suggesting it possesses the broadest knowledge coverage, although its $r$ is lower than that of OpenAI’s GPT-5 and o3. Both GPT-5 and o3, however, yield the highest $g$ values, indicating that these models are more likely to produce correct answers by chance. On the other hand, \textit{Gemma3-27B} shows the lowest $\theta$, indicating a comparatively narrower knowledge base. Meanwhile, \textit{Claude Opus 4.1} shows the lowest $r$ among all models, even below all open-source models listed here, which may stem from its stronger emphasis on code-related reasoning or function-calling tasks rather than visual–language understanding.

\begin{figure}[t]
    \centering
    \captionsetup{skip=3pt}
    \begin{subfigure}[b]{0.67\textwidth}
        \centering
        \captionsetup{skip=1pt}
        \includegraphics[width=\textwidth]{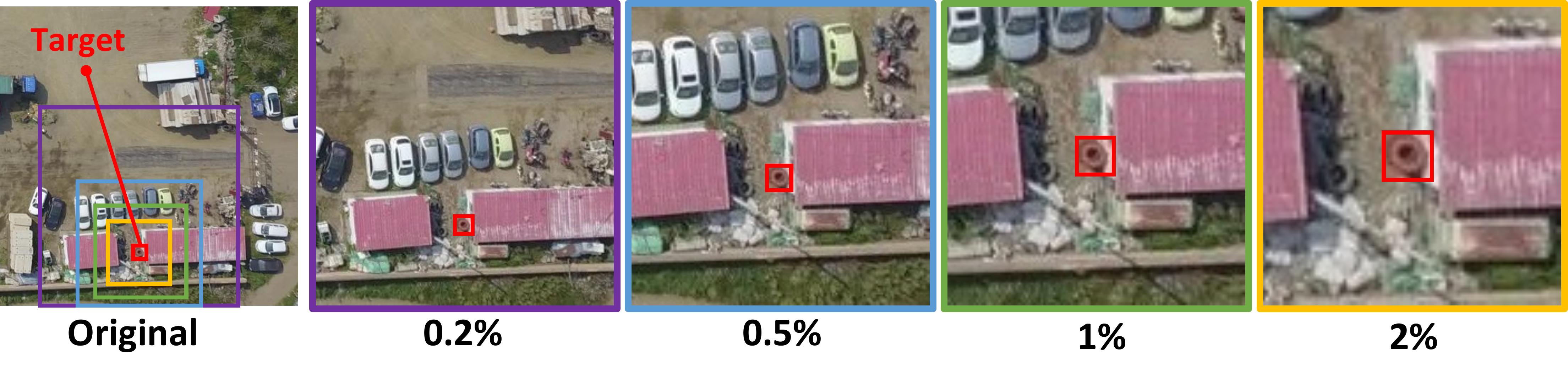}
        \caption{Tire detection at varying magnification levels.}
        \label{fig:zoom_illustration}
    \end{subfigure}
    \hfill
    \begin{subfigure}[b]{0.32\textwidth}
        \centering
        \captionsetup{skip=1pt}
        \includegraphics[width=\textwidth]{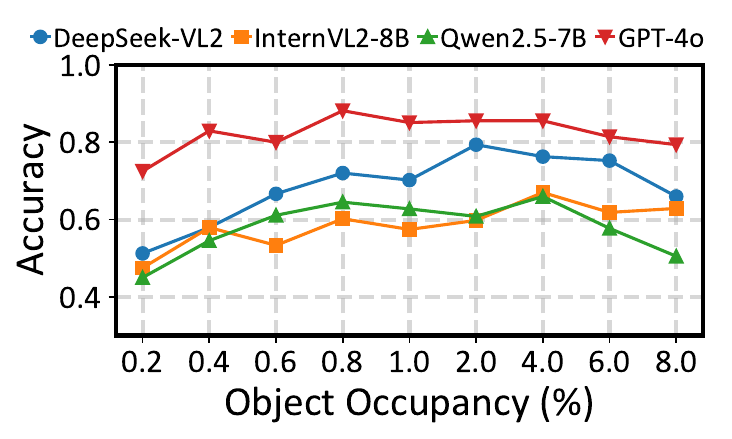}
        \caption{Object size vs. accuracy}
        \label{fig:zoom_performance}
    \end{subfigure}
    \caption{Impact of digital magnification on aerial object detection performance.}
    \label{fig:zoom_in_analysis}
\end{figure}

\section{Case Studies}\label{sec:case_study}
Top-down images are typically captured from high altitudes, which introduces unique challenges such as small object size, unusual perspective, and the lack of depth cues, yet depth is critical for tasks like building height estimation or drone navigation. To examine these challenges, we design four targeted case studies.

\begin{figure}[t]
    \centering
    \captionsetup{skip=3pt}
    \begin{subfigure}[b]{0.58\textwidth}
        \centering
        \captionsetup{skip=1pt}
        \includegraphics[width=\textwidth]{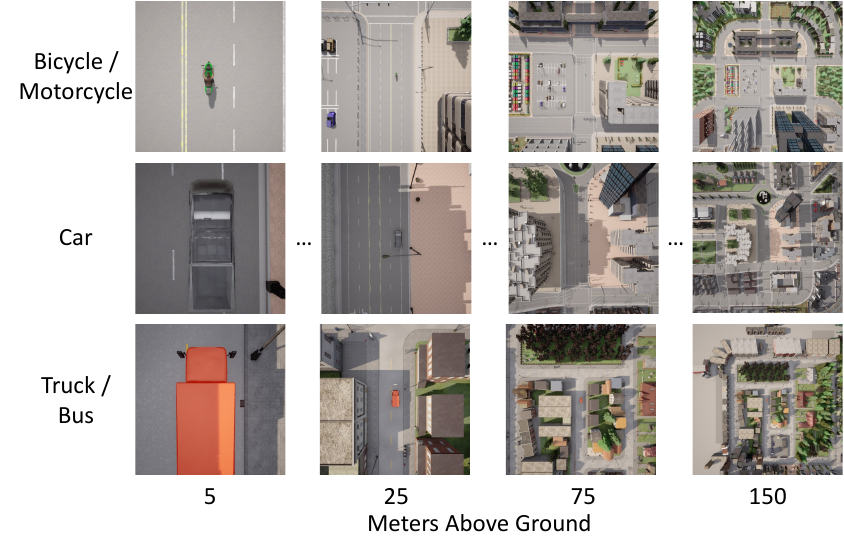}
        \caption{CALRA Simulation for multi-altitude image capture.}
        \label{fig:physical_zoom_illustration}
    \end{subfigure}
    \begin{subfigure}[b]{0.3\textwidth}
        \centering
        \captionsetup{skip=1pt}
        \includegraphics[width=\textwidth]{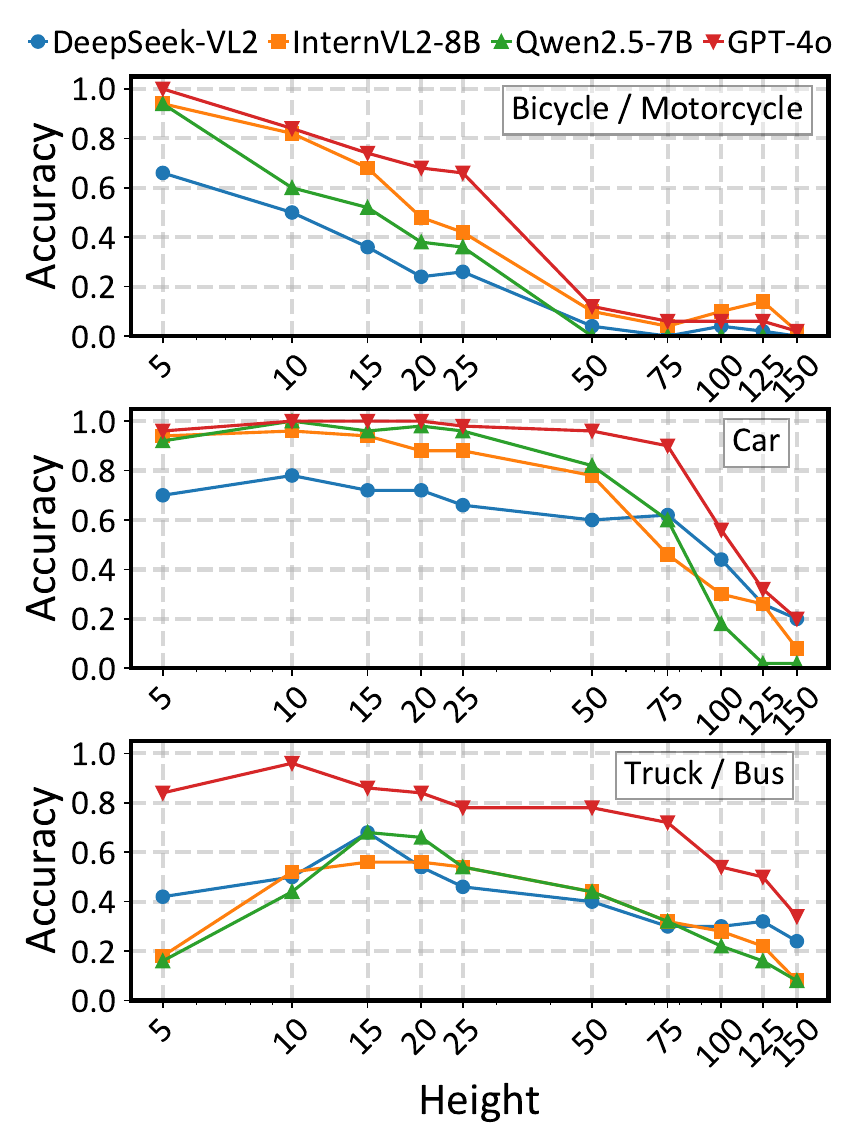}
        \caption{Altitude vs. accuracy.}
        \label{fig:physical_zoom_performance}
    \end{subfigure}
    \caption{Impact of camera altitude on object detection performance. The right plot shows detection accuracy as a function of altitude (5-150m) on a logarithmic scale.}
    \label{fig:physical_zoom_in_analysis}
\end{figure}

\begin{figure}[t]
\centering
\captionsetup{skip=3pt}
\includegraphics[width=0.8\linewidth]{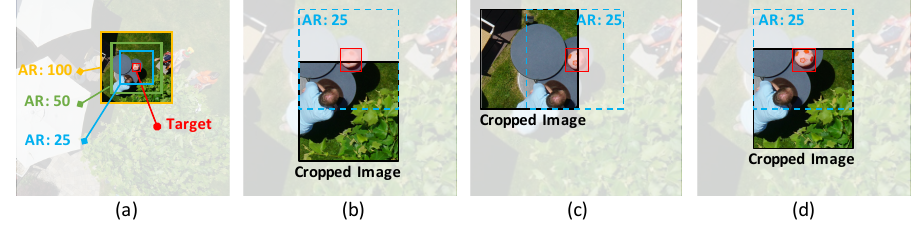}
\caption{Example of Integrity study. (a) Three different area ratios (AR) (25$\times$, 50$\times$, 100$\times$). (b), (c), and (d) show visibility ratios of 30\%, 60\%, and 90\%, respectively, in the setting (AR=25$\times$), depending on how much of the object is bounded inside the image cropped regions.}
\label{fig:integrity}  
\end{figure}

\subsection{Case Study 1: Digital Magnification for Small Object Detection}\label{sec:case1}
Small objects occupy very few pixels, making them difficult for VLMs to detect. We explore a \textit{digital magnification} strategy that crops images to increase the target object’s relative pixel coverage (area ratio), as illustrated in Figure~\ref{fig:zoom_illustration}. We use samples from \textit{object presence} and \textit{object localization} tasks where baseline performance was low, and reformat them using the \textit{object presence} template. 

Figure~\ref{fig:zoom_performance} shows that accuracy rises with area ratio before dropping as context is lost. GPT-4o peaks at only 0.8\% occupancy, whereas open-source models require 2–4\%. Beyond 6\%, performance declines across all models due to resolution loss and reduced context. These findings offer practical guidance on magnification levels for aerial imaging and suggest future work on improving small-object detection in VLMs, particularly for models using multi-tile preprocessing, where tile size could be adapted based on prior knowledge of target object scale.

\subsection{Case Study 2: Altitude Effects on Object Detection}
This study examines optimal hovering heights for drones with a fixed field of view (FOV) when performing tasks that require consistent object detection, such as tracking suspects. Unlike previous studies, we focus on physical ``zoom-in'', where the drone adjusts its altitude to improve detection performance. Because most datasets lack camera height metadata, we used the CARLA simulation to deploy multiple cameras at different altitudes over identical scenes (Figure~\ref{fig:physical_zoom_illustration}). We evaluated three object categories (bicycle/motorcycle, car, and truck/bus)—chosen for their frequency in aerial tasks and distinct size differences. \textit{Object presence performance} was measured across altitudes from 5 to 150 meters, spanning typical operational ranges for commercial and tactical drones, while keeping image resolution constant. This setup offers practical guidance for maximizing detection reliability through optimal drone positioning rather than post-capture image processing.

As shown in Figure~\ref{fig:physical_zoom_performance}, accuracy generally decreases with altitude but peaks at specific heights: 5m for bicycles/motorcycles, 10m for cars, and 15m for trucks/buses. We attribute this to field coverage differences: at low altitudes, large objects may be only partially visible, reducing detection accuracy, while smaller objects remain fully visible even at minimal heights.

\subsection{Case Study 3: Object Visibility and Partial Occlusion}
Objects may be only partially visible, especially near image borders. We controlled visibility (\textbf{integrity}) by shifting a fixed-size crop window over objects at a set area ratio (AR) (Figure~\ref{fig:integrity}). This allowed us to vary integrity while keeping magnification constant.

\begin{figure}[t]
    \centering
    \captionsetup{skip=3pt}
    \begin{subfigure}[b]{0.73\textwidth}
        \centering
        \captionsetup{skip=1pt}
        \includegraphics[width=\textwidth]{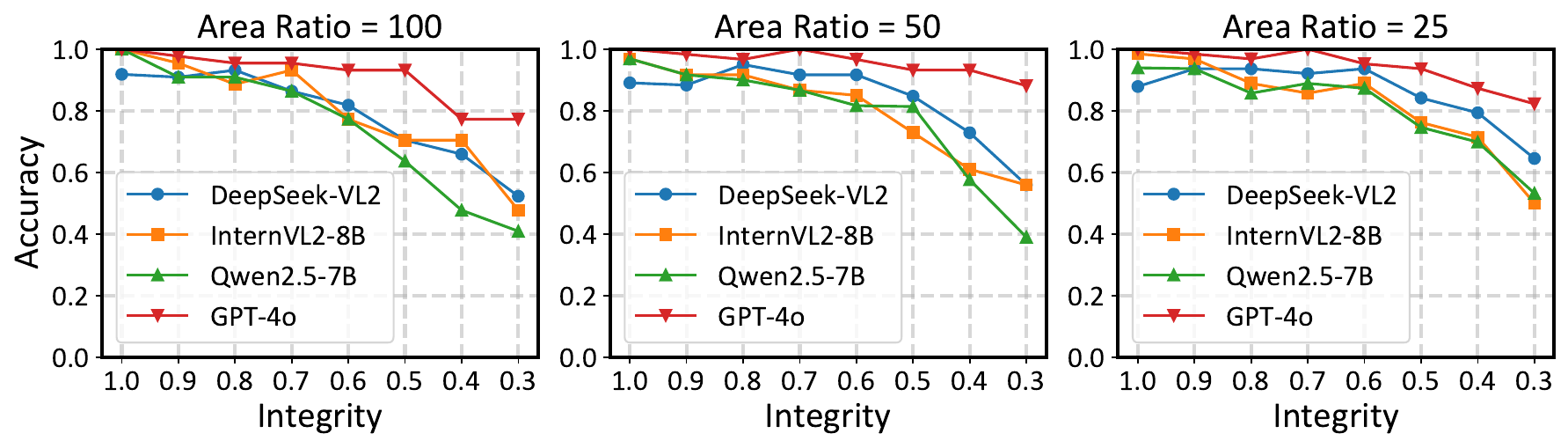}
        \caption{Object Integrity vs. Model Accuracy under three Area Ratio Settings.}
        \label{fig:integrity_results}
    \end{subfigure}
    \hfill
    \begin{subfigure}[b]{0.25\textwidth}
        \centering
        \captionsetup{skip=1pt}
        \includegraphics[width=\textwidth]{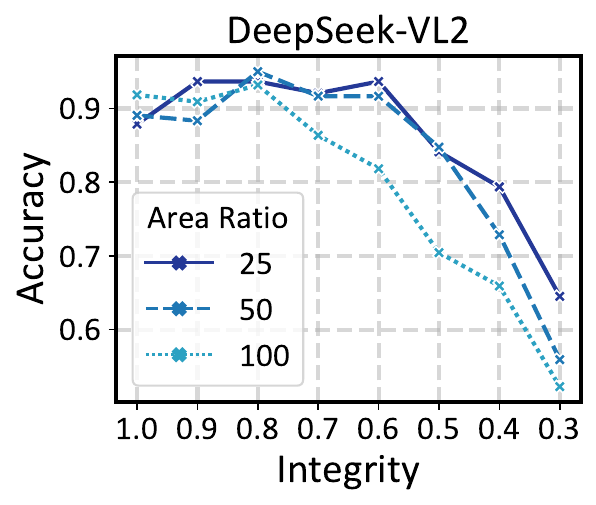}
        \caption{AR-integrity.}
        \label{fig:AR_completeness}
    \end{subfigure}
    \caption{Impact of Object Integrity and Area Ratio on VLM Performance.}
    \label{fig:integrity_study}
\end{figure}

Figure~\ref{fig:integrity_results} shows that accuracy stays stable ($\geq$90\%) until integrity drops below a threshold, then declines sharply. 
This threshold depends on AR: with AR=100, accuracy drops below 70\% integrity, while lower ARs fail around 60\%(Figure~\ref{fig:AR_completeness}). This demonstrates how incomplete visibility affects detection even without resolution changes.

\subsection{Case Study 4: Z-Axis Perception and Depth Understanding}
\begin{figure}[t]
    \centering
    \captionsetup{skip=3pt}
    \begin{subfigure}[b]{0.43\textwidth}
        \centering
        \includegraphics[width=\textwidth]{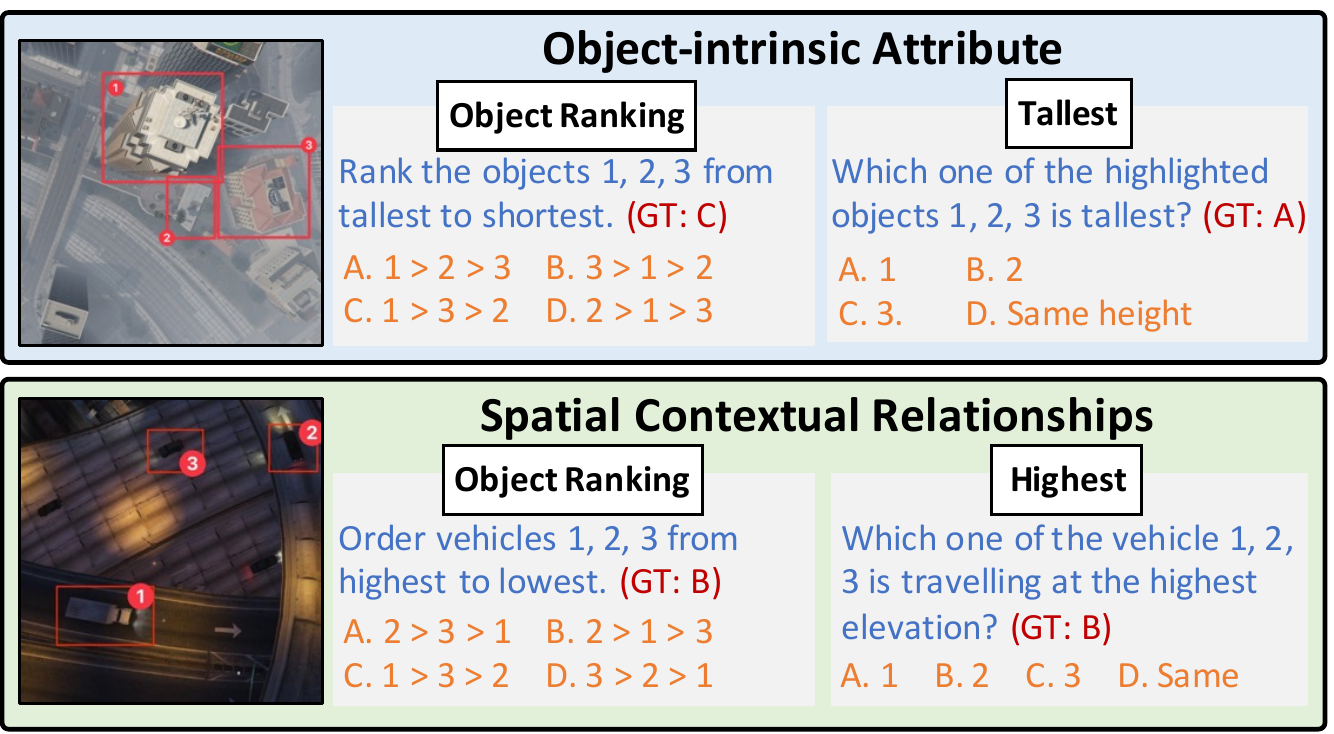}
        \caption{Example of two types of questions.}
        \label{fig:height_mcq_rank}
    \end{subfigure}
    \hfill
    \begin{subfigure}[b]{0.48\textwidth}
        \centering
         \includegraphics[width=\textwidth]{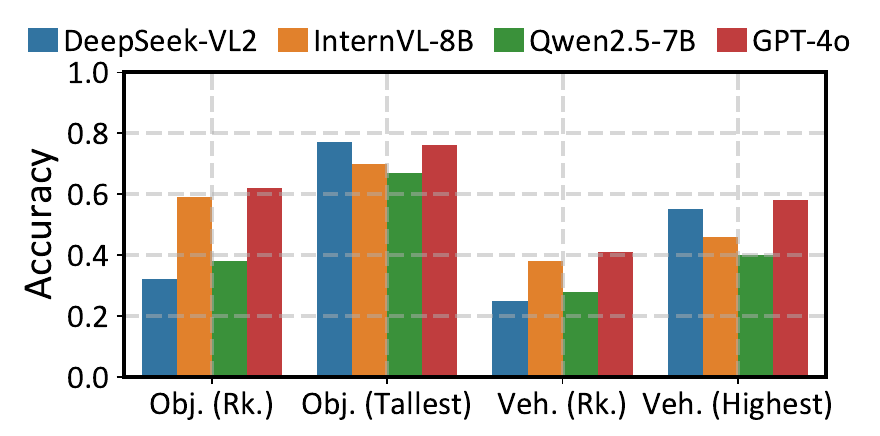}
        \caption{Depth Performance.}
        \label{fig:z_eval}
    \end{subfigure}
    \caption{Analysis of spatial awareness and depth perception.}
    \label{fig:depth_analysis}
    \vspace{-0.2in}
\end{figure}

Since top-down images preserve xy-plane information, they inherently lack altitude cues. To evaluate this limitation, we defined two types of \textbf{z-axis awareness} challenges (Figure~\ref{fig:height_mcq_rank}): (i) assessing an object's intrinsic properties, such as a building’s or tree’s height, and (ii) evaluating contextual relationships, such as determining whether a car is traveling on a road or an overpass. As shown in Figure~\ref{fig:z_eval}, DeepSeek performs well on tallest/highest identification but struggles with ranking tasks, whereas GPT-4o achieves near-best performance across both types.

\section{Discussion}
\paragraph{Probing Intrinsic Model Properties}
Although we introduce these metrics within the context of \pname, they should not be viewed as entirely independent of the dataset. Rather, \pname provides a controlled medium to probe latent aspects of model behavior that are difficult to observe otherwise. The estimated parameters $(\theta,r,g)$ offer a useful abstraction of how much correctness stems from genuine knowledge versus chance, but their validity ultimately depends on the diversity and fidelity of the underlying data. Future work should compare these estimates across multiple datasets and domains to confirm whether they generalize as truly model-intrinsic properties.
\paragraph{Difference with Multi-Pass Evaluation and Majority Voting}
Unlike multi-pass evaluation with majority voting, which explores variability in stochastic decoding, our RotationalEval (RE) assesses invariance under controlled changes to the visual input. This makes RE more diagnostic of grounding and invariance. At the same time, RE introduces stricter success criteria, which may amplify penalties for minor inconsistencies or ambiguous visual cases. Thus, we view RE as complementary to multi-pass evaluation: the former highlights robustness to perceptual transformations, while the latter captures robustness to sampling variance.
\paragraph{Limitations of \pname}
Our work has several limitations. (1) The current benchmark includes 2,000 core questions and additional case-study questions, which, although carefully curated, remain relatively small compared to large-scale multimodal datasets. We are working to expand both the size and diversity of the benchmark. (2) Except for visual grounding tasks, correctness of generated responses is determined by an extractor model, and our reported results rely on GPT-4o. This introduces possible bias and raises the need for extractor-agnostic evaluation. (3) A significant portion of case studies in the benchmark is based on simulation, which allows precise control but may not fully capture the variability of real-world aerial images. Future iterations should incorporate larger amounts of real data. 

\section{Conclusion}
In this work, we introduced \pname, a comprehensive benchmark for evaluating VLMs on top-down images, comprising over 2,000 manually labeled questions across diverse categories. To ensure robust and reliable assessment, we proposed \textbf{RotationalEval}, an evaluation strategy that leverages the rotational invariance of top-down perspectives to provide a more rigorous alternative to standard single-pass evaluation. Beyond accuracy, we further developed a set of \textbf{reliability-oriented metrics} that assess how much of a model’s performance stems from genuine knowledge rather than lucky guesses or hallucinated responses. Our multi-dimensional analysis reveals both the capabilities and limitations of current VLMs, and our four case studies demonstrate their strengths and challenges in real-world aerial applications. While \pname serves as the testbed for this study, these metrics are not tied to any specific dataset and can serve as \textbf{general probes of model reliability}, offering a new perspective for guiding future development of more trustworthy VLMs.

\bibliography{reference}

\begin{thebibliography}{52}
\providecommand{\natexlab}[1]{#1}
\providecommand{\url}[1]{\texttt{#1}}
\expandafter\ifx\csname urlstyle\endcsname\relax
  \providecommand{\doi}[1]{doi: #1}\else
  \providecommand{\doi}{doi: \begingroup \urlstyle{rm}\Url}\fi

\bibitem[Anthropic(2024)]{anthropic2024claude35sonnet}
Anthropic.
\newblock Claude 3.5 sonnet, 2024.
\newblock URL \url{https://www.anthropic.com/news/claude-3-5-sonnet}.
\newblock Accessed: 2025-03-26.

\bibitem[Anthropic(2025{\natexlab{a}})]{anthropic2024claude37sonnet}
Anthropic.
\newblock Claude 3.7 sonnet, 2025{\natexlab{a}}.
\newblock URL \url{https://www.anthropic.com/news/claude-3-7-sonnet}.
\newblock Accessed: 2025-09-16.

\bibitem[Anthropic(2025{\natexlab{b}})]{anthropic2024claude41opus}
Anthropic.
\newblock Claude 4.1, 2025{\natexlab{b}}.
\newblock URL \url{https://www.anthropic.com/news/claude-opus-4-1}.
\newblock Accessed: 2025-09-16.

\bibitem[Anthropic(2025{\natexlab{c}})]{anthropic2024claude4sonnet}
Anthropic.
\newblock Claude 4, 2025{\natexlab{c}}.
\newblock URL \url{https://www.anthropic.com/news/claude-4}.
\newblock Accessed: 2025-09-16.

\bibitem[Bai et~al.(2025{\natexlab{a}})Bai, Chen, and et~al]{Qwen2.5-VL}
Shuai Bai, Keqin Chen, and Xuejing~Liu et~al.
\newblock Qwen2.5-vl technical report.
\newblock \emph{arXiv preprint arXiv:2502.13923}, 2025{\natexlab{a}}.

\bibitem[Bai et~al.(2025{\natexlab{b}})Bai, Wang, Xiao, He, Han, Zhang, and Shou]{bai2025hallucination}
Zechen Bai, Pichao Wang, Tianjun Xiao, Tong He, Zongbo Han, Zheng Zhang, and Mike~Zheng Shou.
\newblock Hallucination of multimodal large language models: A survey, 2025{\natexlab{b}}.
\newblock URL \url{https://arxiv.org/abs/2404.18930}.

\bibitem[Chen et~al.(2025)Chen, Wang, Cao, Liu, Gao, et~al.]{chen2025internvl25}
Zhe Chen, Weiyun Wang, Yue Cao, Yangzhou Liu, Zhangwei Gao, et~al.
\newblock Expanding performance boundaries of open-source multimodal models with model, data, and test-time scaling, 2025.
\newblock URL \url{https://arxiv.org/abs/2412.05271}.

\bibitem[Danish et~al.(2025)Danish, Munir, Shah, Kuckreja, Khan, Fraccaro, Lacoste, and Khan]{danish2025geobench}
Muhammad~Sohail Danish, Muhammad~Akhtar Munir, Syed Roshaan~Ali Shah, Kartik Kuckreja, Fahad~Shahbaz Khan, Paolo Fraccaro, Alexandre Lacoste, and Salman Khan.
\newblock Geobench-vlm: Benchmarking vision-language models for geospatial tasks, 2025.
\newblock URL \url{https://arxiv.org/abs/2411.19325}.

\bibitem[Dosovitskiy et~al.(2017)Dosovitskiy, Ros, Codevilla, Lopez, and Koltun]{Dosovitskiy17}
Alexey Dosovitskiy, German Ros, Felipe Codevilla, Antonio Lopez, and Vladlen Koltun.
\newblock {CARLA}: {An} open urban driving simulator.
\newblock In \emph{Proceedings of the 1st Annual Conference on Robot Learning}, pp.\  1--16, 2017.

\bibitem[Duan et~al.(2024)Duan, Yang, Qiao, Fang, Chen, Liu, Dong, Zang, Zhang, Wang, et~al.]{duan2024vlmevalkit}
Haodong Duan, Junming Yang, Yuxuan Qiao, Xinyu Fang, Lin Chen, Yuan Liu, Xiaoyi Dong, Yuhang Zang, Pan Zhang, Jiaqi Wang, et~al.
\newblock Vlmevalkit: An open-source toolkit for evaluating large multi-modality models.
\newblock In \emph{Proceedings of the 32nd ACM International Conference on Multimedia}, pp.\  11198--11201, 2024.

\bibitem[Fu et~al.(2024)Fu, Chen, Shen, Qin, Zhang, Lin, Yang, Zheng, Li, Sun, Wu, and Ji]{fu2024mme}
Chaoyou Fu, Peixian Chen, Yunhang Shen, Yulei Qin, Mengdan Zhang, Xu~Lin, Jinrui Yang, Xiawu Zheng, Ke~Li, Xing Sun, Yunsheng Wu, and Rongrong Ji.
\newblock Mme: A comprehensive evaluation benchmark for multimodal large language models, 2024.
\newblock URL \url{https://arxiv.org/abs/2306.13394}.

\bibitem[Gasienica-Jozkowy et~al.(2021)Gasienica-Jozkowy, Knapik, and Cyganek]{gasienica2021ensemble}
Jan Gasienica-Jozkowy, Mateusz Knapik, and Boguslaw Cyganek.
\newblock An ensemble deep learning method with optimized weights for drone-based water rescue and surveillance.
\newblock \emph{Integrated Computer-Aided Engineering}, 28\penalty0 (1):\penalty0 1--15, 01 2021.
\newblock \doi{10.3233/ICA-210649}.

\bibitem[Google(2024)]{geminiteam2024gemini}
Google.
\newblock Gemini 1.5: Unlocking multimodal understanding across millions of tokens of context, 2024.
\newblock URL \url{https://arxiv.org/abs/2403.05530}.

\bibitem[Google(2025{\natexlab{a}})]{geminiteam2025gemini}
Google.
\newblock Gemini 2.5: Pushing the frontier with advanced reasoning, multimodality, long context, and next generation agentic capabilities, 2025{\natexlab{a}}.
\newblock URL \url{https://arxiv.org/abs/2507.06261}.

\bibitem[Google(2025{\natexlab{b}})]{gemmateam2025gemma3technicalreport}
Google.
\newblock Gemma 3 technical report, 2025{\natexlab{b}}.
\newblock URL \url{https://arxiv.org/abs/2503.19786}.

\bibitem[Guan et~al.(2024)Guan, Liu, Wu, Xian, Li, Liu, Wang, Chen, Huang, Yacoob, Manocha, and Zhou]{guan2024hallusionbench}
Tianrui Guan, Fuxiao Liu, Xiyang Wu, Ruiqi Xian, Zongxia Li, Xiaoyu Liu, Xijun Wang, Lichang Chen, Furong Huang, Yaser Yacoob, Dinesh Manocha, and Tianyi Zhou.
\newblock Hallusionbench: An advanced diagnostic suite for entangled language hallucination and visual illusion in large vision-language models, 2024.
\newblock URL \url{https://arxiv.org/abs/2310.14566}.

\bibitem[Hu et~al.(2023)Hu, Yuan, Wen, Lu, and Li]{hu2023rsgpt}
Yuan Hu, Jianlong Yuan, Congcong Wen, Xiaonan Lu, and Xiang Li.
\newblock Rsgpt: A remote sensing vision language model and benchmark, 2023.
\newblock URL \url{https://arxiv.org/abs/2307.15266}.

\bibitem[{ICG}(2019)]{tugraz2025drone}
{ICG}.
\newblock Semantic segmentation drone dataset, 2019.
\newblock URL \url{http://dronedataset.icg.tugraz.at/}.

\bibitem[KimiTeam(2025)]{kimiteam2025kimivltechnicalreport}
KimiTeam.
\newblock Kimi-vl technical report, 2025.
\newblock URL \url{https://arxiv.org/abs/2504.07491}.

\bibitem[Kuckreja et~al.(2023)Kuckreja, Danish, Naseer, Das, Khan, and Khan]{kuckreja2023geochat}
Kartik Kuckreja, Muhammad~Sohail Danish, Muzammal Naseer, Abhijit Das, Salman Khan, and Fahad~Shahbaz Khan.
\newblock Geochat: Grounded large vision-language model for remote sensing, 2023.
\newblock URL \url{https://arxiv.org/abs/2311.15826}.

\bibitem[Li et~al.(2024{\natexlab{a}})Li, Zhang, Zhou, Collier, Korhonen, and Vulić]{li2024topviewrs}
Chengzu Li, Caiqi Zhang, Han Zhou, Nigel Collier, Anna Korhonen, and Ivan Vulić.
\newblock Topviewrs: Vision-language models as top-view spatial reasoners, 2024{\natexlab{a}}.
\newblock URL \url{https://arxiv.org/abs/2406.02537}.

\bibitem[Li et~al.(2024{\natexlab{b}})Li, Zhang, Zhang, Zhang, Li, Li, Ma, and Li]{li2024llava}
Feng Li, Renrui Zhang, Hao Zhang, Yuanhan Zhang, Bo~Li, Wei Li, Zejun Ma, and Chunyuan Li.
\newblock Llava-next-interleave: Tackling multi-image, video, and 3d in large multimodal models.
\newblock \emph{arXiv preprint arXiv:2407.07895}, 2024{\natexlab{b}}.

\bibitem[Li et~al.(2023{\natexlab{a}})Li, Du, Zhou, Wang, Zhao, and Wen]{li2023evaluating}
Yifan Li, Yifan Du, Kun Zhou, Jinpeng Wang, Wayne~Xin Zhao, and Ji-Rong Wen.
\newblock Evaluating object hallucination in large vision-language models, 2023{\natexlab{a}}.
\newblock URL \url{https://arxiv.org/abs/2305.10355}.

\bibitem[Li et~al.(2023{\natexlab{b}})Li, Du, Zhou, Wang, Zhao, and Wen]{li2023hallucination}
Yifan Li, Yifan Du, Kun Zhou, Jinpeng Wang, Xin Zhao, and Ji-Rong Wen.
\newblock Evaluating object hallucination in large vision-language models.
\newblock In Houda Bouamor, Juan Pino, and Kalika Bali (eds.), \emph{Proceedings of the 2023 Conference on Empirical Methods in Natural Language Processing}, pp.\  292--305, Singapore, December 2023{\natexlab{b}}. Association for Computational Linguistics.
\newblock \doi{10.18653/v1/2023.emnlp-main.20}.
\newblock URL \url{https://aclanthology.org/2023.emnlp-main.20/}.

\bibitem[Lin et~al.(2015)Lin, Maire, Belongie, Bourdev, Girshick, Hays, Perona, Ramanan, Zitnick, and Dollár]{lin2015coco}
Tsung-Yi Lin, Michael Maire, Serge Belongie, Lubomir Bourdev, Ross Girshick, James Hays, Pietro Perona, Deva Ramanan, C.~Lawrence Zitnick, and Piotr Dollár.
\newblock Microsoft coco: Common objects in context, 2015.
\newblock URL \url{https://arxiv.org/abs/1405.0312}.

\bibitem[Liu et~al.(2024{\natexlab{a}})Liu, Xue, Chen, Chen, Zhao, Wang, Hou, Li, and Peng]{liu2024hallucination}
Hanchao Liu, Wenyuan Xue, Yifei Chen, Dapeng Chen, Xiutian Zhao, Ke~Wang, Liping Hou, Rongjun Li, and Wei Peng.
\newblock A survey on hallucination in large vision-language models, 2024{\natexlab{a}}.
\newblock URL \url{https://arxiv.org/abs/2402.00253}.

\bibitem[Liu et~al.(2023)Liu, Li, Wu, and Lee]{liu2023llava}
Haotian Liu, Chunyuan Li, Qingyang Wu, and Yong~Jae Lee.
\newblock Visual instruction tuning.
\newblock In \emph{NeurIPS}, 2023.

\bibitem[Liu et~al.(2024{\natexlab{b}})Liu, Duan, Zhang, Li, Zhang, Zhao, Yuan, Wang, He, Liu, Chen, and Lin]{liu2024mmbench}
Yuan Liu, Haodong Duan, Yuanhan Zhang, Bo~Li, Songyang Zhang, Wangbo Zhao, Yike Yuan, Jiaqi Wang, Conghui He, Ziwei Liu, Kai Chen, and Dahua Lin.
\newblock Mmbench: Is your multi-modal model an all-around player?, 2024{\natexlab{b}}.
\newblock URL \url{https://arxiv.org/abs/2307.06281}.

\bibitem[Lu et~al.(2024)Lu, Bansal, Xia, Liu, Li, Hajishirzi, Cheng, Chang, Galley, and Gao]{lu2024mathvista}
Pan Lu, Hritik Bansal, Tony Xia, Jiacheng Liu, Chunyuan Li, Hannaneh Hajishirzi, Hao Cheng, Kai-Wei Chang, Michel Galley, and Jianfeng Gao.
\newblock Mathvista: Evaluating mathematical reasoning of foundation models in visual contexts, 2024.
\newblock URL \url{https://arxiv.org/abs/2310.02255}.

\bibitem[Lu et~al.(2018)Lu, Xue, Xia, and Zhang]{Lu2018UAV}
Yuncheng Lu, Zhucun Xue, Gui-Song Xia, and Liangpei Zhang.
\newblock A survey on vision-based uav navigation.
\newblock \emph{Geo-spatial Information Science}, 21\penalty0 (1):\penalty0 21--32, 2018.
\newblock \doi{10.1080/10095020.2017.1420509}.
\newblock URL \url{https://doi.org/10.1080/10095020.2017.1420509}.

\bibitem[Mou et~al.(in press)Mou, Hua, Jin, and Zhu]{eradataset}
L.~Mou, Y.~Hua, P.~Jin, and X.~X. Zhu.
\newblock {ERA: A dataset and deep learning benchmark for event recognition in aerial videos}.
\newblock \emph{IEEE Geoscience and Remote Sensing Magazine}, in press.

\bibitem[Muhtar et~al.(2024)Muhtar, Li, Gu, Zhang, and Xiao]{muhtar2024lhrs}
Dilxat Muhtar, Zhenshi Li, Feng Gu, Xueliang Zhang, and Pengfeng Xiao.
\newblock Lhrs-bot: Empowering remote sensing with vgi-enhanced large multimodal language model, 2024.
\newblock URL \url{https://arxiv.org/abs/2402.02544}.

\bibitem[Nearmap(2022)]{Nearmap2022aerial}
Nearmap.
\newblock How aerial imagery revolutionizes urban planning, 2022.
\newblock URL \url{https://www.nearmap.com/blog/how-aerial-imagery-revolutionizes-urban-planning}.
\newblock Accessed: 2025-09-16.

\bibitem[OpenAI(2024)]{openai2024gpt4}
OpenAI.
\newblock Gpt-4 technical report, 2024.
\newblock URL \url{https://arxiv.org/abs/2303.08774}.

\bibitem[OpenAI(2025{\natexlab{a}})]{openai2025gpt41}
OpenAI.
\newblock Introducing gpt-4.1 in the api, 2025{\natexlab{a}}.
\newblock URL \url{https://openai.com/index/gpt-4-1/}.
\newblock Accessed: 2025-09-16.

\bibitem[OpenAI(2025{\natexlab{b}})]{openai2025gpt5}
OpenAI.
\newblock Introducing gpt-5, 2025{\natexlab{b}}.
\newblock URL \url{https://openai.com/index/introducing-gpt-5/}.
\newblock Accessed: 2025-09-16.

\bibitem[OpenAI(2025{\natexlab{c}})]{openai2025gpto3}
OpenAI.
\newblock Introducing openai o3 and o4-mini, 2025{\natexlab{c}}.
\newblock URL \url{https://openai.com/index/introducing-o3-and-o4-mini/}.
\newblock Accessed: 2025-09-16.

\bibitem[Radford et~al.(2021)Radford, Kim, Hallacy, Ramesh, Goh, Agarwal, Sastry, Askell, Mishkin, Clark, Krueger, and Sutskever]{radford2021clip}
Alec Radford, Jong~Wook Kim, Chris Hallacy, Aditya Ramesh, Gabriel Goh, Sandhini Agarwal, Girish Sastry, Amanda Askell, Pamela Mishkin, Jack Clark, Gretchen Krueger, and Ilya Sutskever.
\newblock Learning transferable visual models from natural language supervision, 2021.
\newblock URL \url{https://arxiv.org/abs/2103.00020}.

\bibitem[Russakovsky et~al.(2015)Russakovsky, Deng, Su, Krause, Satheesh, Ma, Huang, Karpathy, Khosla, Bernstein, Berg, and Fei-Fei]{russakovsky2015imagenet}
Olga Russakovsky, Jia Deng, Hao Su, Jonathan Krause, Sanjeev Satheesh, Sean Ma, Zhiheng Huang, Andrej Karpathy, Aditya Khosla, Michael Bernstein, Alexander~C. Berg, and Li~Fei-Fei.
\newblock Imagenet large scale visual recognition challenge, 2015.
\newblock URL \url{https://arxiv.org/abs/1409.0575}.

\bibitem[Schuhmann et~al.(2022)Schuhmann, Beaumont, Vencu, Gordon, Wightman, Cherti, Coombes, Katta, Mullis, Wortsman, Schramowski, Kundurthy, Crowson, Schmidt, Kaczmarczyk, and Jitsev]{schuhmann2022laion5b}
Christoph Schuhmann, Romain Beaumont, Richard Vencu, Cade Gordon, Ross Wightman, Mehdi Cherti, Theo Coombes, Aarush Katta, Clayton Mullis, Mitchell Wortsman, Patrick Schramowski, Srivatsa Kundurthy, Katherine Crowson, Ludwig Schmidt, Robert Kaczmarczyk, and Jenia Jitsev.
\newblock Laion-5b: An open large-scale dataset for training next generation image-text models, 2022.
\newblock URL \url{https://arxiv.org/abs/2210.08402}.

\bibitem[Shaha(2025)]{roboflow2025shaha}
Shaha.
\newblock Aerial traffic images.
\newblock \url{https://universe.roboflow.com/cg-0fmsf/shaha-adfy7}, 2025.
\newblock Accessed: 2025-03-26.

\bibitem[Shen et~al.(2025)Shen, Liu, Li, Fang, Ma, Liao, Shen, Zhang, Zhao, Zhang, Xu, and Zhao]{shen2025vlmr1stablegeneralizabler1style}
Haozhan Shen, Peng Liu, Jingcheng Li, Chunxin Fang, Yibo Ma, Jiajia Liao, Qiaoli Shen, Zilun Zhang, Kangjia Zhao, Qianqian Zhang, Ruochen Xu, and Tiancheng Zhao.
\newblock Vlm-r1: A stable and generalizable r1-style large vision-language model, 2025.
\newblock URL \url{https://arxiv.org/abs/2504.07615}.

\bibitem[Varga et~al.(2022)Varga, Kiefer, Messmer, and Zell]{varga2022seadronessee}
Leon~Amadeus Varga, Benjamin Kiefer, Martin Messmer, and Andreas Zell.
\newblock Seadronessee: A maritime benchmark for detecting humans in open water.
\newblock In \emph{Proceedings of the IEEE/CVF Winter Conference on Applications of Computer Vision}, pp.\  2260--2270, 2022.

\bibitem[Wang et~al.(2024)Wang, Chen, Zhang, Tian, Xu, Deng, and Chen]{wang2024mllmseedynamiccorrection}
Chenxi Wang, Xiang Chen, Ningyu Zhang, Bozhong Tian, Haoming Xu, Shumin Deng, and Huajun Chen.
\newblock Mllm can see? dynamic correction decoding for hallucination mitigation, 2024.
\newblock URL \url{https://arxiv.org/abs/2410.11779}.

\bibitem[Wang et~al.(2025)Wang, Gao, Gu, Pu, et~al.]{wang2025internvl35}
Weiyun Wang, Zhangwei Gao, Lixin Gu, Hengjun Pu, et~al.
\newblock Internvl3.5: Advancing open-source multimodal models in versatility, reasoning, and efficiency, 2025.
\newblock URL \url{https://arxiv.org/abs/2508.18265}.

\bibitem[Wu et~al.(2024)Wu, Chen, Pan, et~al.]{wu2024deepseekvl2}
Zhiyu Wu, Xiaokang Chen, Zizheng Pan, et~al.
\newblock Deepseek-vl2: Mixture-of-experts vision-language models for advanced multimodal understanding, 2024.
\newblock URL \url{https://arxiv.org/abs/2412.10302}.

\bibitem[Yu et~al.(2024)Yu, Yang, Li, Wang, Lin, Liu, Wang, and Wang]{yu2024mmvet}
Weihao Yu, Zhengyuan Yang, Linjie Li, Jianfeng Wang, Kevin Lin, Zicheng Liu, Xinchao Wang, and Lijuan Wang.
\newblock Mm-vet: Evaluating large multimodal models for integrated capabilities, 2024.
\newblock URL \url{https://arxiv.org/abs/2308.02490}.

\bibitem[Yue et~al.(2024)Yue, Ni, Zhang, Zheng, Liu, Zhang, Stevens, Jiang, Ren, Sun, Wei, Yu, Yuan, Sun, Yin, Zheng, Yang, Liu, Huang, Sun, Su, and Chen]{yue2024mmmu}
Xiang Yue, Yuansheng Ni, Kai Zhang, Tianyu Zheng, Ruoqi Liu, Ge~Zhang, Samuel Stevens, Dongfu Jiang, Weiming Ren, Yuxuan Sun, Cong Wei, Botao Yu, Ruibin Yuan, Renliang Sun, Ming Yin, Boyuan Zheng, Zhenzhu Yang, Yibo Liu, Wenhao Huang, Huan Sun, Yu~Su, and Wenhu Chen.
\newblock Mmmu: A massive multi-discipline multimodal understanding and reasoning benchmark for expert agi, 2024.
\newblock URL \url{https://arxiv.org/abs/2311.16502}.

\bibitem[Zhai et~al.(2023)Zhai, Mustafa, Kolesnikov, and Beyer]{zhai2023siglip}
Xiaohua Zhai, Basil Mustafa, Alexander Kolesnikov, and Lucas Beyer.
\newblock Sigmoid loss for language image pre-training, 2023.
\newblock URL \url{https://arxiv.org/abs/2303.15343}.

\bibitem[Zhao et~al.(2025)Zhao, Xia, Hou, Liu, Xia, and Jiang]{zhao2025flexifly}
Minghui Zhao, Junxi Xia, Kaiyuan Hou, Yanchen Liu, Stephen Xia, and Xiaofan Jiang.
\newblock \emph{FlexiFly: Interfacing the Physical World with Foundation Models Empowered by Reconfigurable Drone Systems}, pp.\  463–476.
\newblock Association for Computing Machinery, New York, NY, USA, 2025.
\newblock ISBN 9798400714795.
\newblock URL \url{https://doi.org/10.1145/3715014.3722081}.

\bibitem[Zhu et~al.(2025)Zhu, Wang, Chen, Liu, Ye, et~al.]{zhu2025internvl3}
Jinguo Zhu, Weiyun Wang, Zhe Chen, Zhaoyang Liu, Shenglong Ye, et~al.
\newblock Internvl3: Exploring advanced training and test-time recipes for open-source multimodal models, 2025.
\newblock URL \url{https://arxiv.org/abs/2504.10479}.

\bibitem[Zhu et~al.(2021)Zhu, Wen, Du, Bian, Fan, Hu, and Ling]{zhu2021visdrone}
Pengfei Zhu, Longyin Wen, Dawei Du, Xiao Bian, Heng Fan, Qinghua Hu, and Haibin Ling.
\newblock Detection and tracking meet drones challenge.
\newblock \emph{IEEE Transactions on Pattern Analysis and Machine Intelligence}, 44\penalty0 (11):\penalty0 7380--7399, 2021.

\end{thebibliography}
\bibliographystyle{iclr2026_conference}

\newpage
\appendix
\begin{center}
    \Large\textbf{Appendix}
\end{center}

\section{Identifiability of the Mixture Parameters}\label{app:identifiability}

We used three parameters, $(\theta,r,g)\in[0,1]^3$ to study the reliability of the models. These parameters denote the proportion of questions a model truly knows ($\theta$), model's accuracy among the questions that it knows ($r$), and model's accuracy among the questions it does not know and guessed ($g$).

We show that these parameters in our mixture model are \emph{generically unique} given the observed statistics
\[
\mathrm{RE},\qquad \overline{\mathrm{VE}},\qquad \mathrm{MA}.
\]

\subsection{Problem formulation}

Assume the parameters satisfy
\begin{equation}
\label{eq:mixture_system_app}
\begin{aligned}
\mathrm{RE} &= \theta r^4 + (1-\theta) g^4, \\
\overline{\mathrm{VE}} &= \theta r + (1-\theta) g, \\
\mathrm{MA} &= \theta (1-r)^4 + (1-\theta)(1-g)^4.
\end{aligned}
\end{equation}
This system is symmetric under the transformation
\[
(\theta,r,g)\longleftrightarrow (1-\theta,g,r).
\]
To remove this trivial multiplicity, we restrict to the \emph{ordered domain}
\begin{equation}
\label{eq:ordered_domain_app}
\mathcal{D}_{\overline{\mathrm{VE}}} \;:=\; \{(r,g)\mid 0\le g < \overline{\mathrm{VE}} < r \le 1\},\qquad
\theta \;=\; \frac{\overline{\mathrm{VE}}-g}{r-g}\in(0,1).
\end{equation}
The degenerate case $r=g$ occurs iff $\mathrm{RE}=\overline{\mathrm{VE}}^4$ and $\mathrm{MA}=(1-\overline{\mathrm{VE}})^4$ and is excluded. We also exclude trivial boundary cases $\overline{\mathrm{VE}}\in\{0,1\}$ or $\mathrm{MA}\in\{0,1\}$, where conditioning becomes ill-defined.

\subsection{Reduction to secant equations}

Define $f(x)=x^4$ and $u(x)=(1-x)^4$. Eliminating $\theta$ using the middle equation in \eqref{eq:mixture_system_app}, the outer equations become the \emph{secant identities}
\begin{equation}
\label{eq:secant_form_app}
\boxed{\ 
\frac{\mathrm{RE}-g^4}{\,\overline{\mathrm{VE}}-g\,}=\frac{r^4-g^4}{\,r-g\,},\qquad
\frac{\mathrm{MA}-(1-g)^4}{\,\overline{\mathrm{VE}}-g\,}=\frac{(1-r)^4-(1-g)^4}{\,r-g\,}}\ .
\end{equation}
These state that $(g,r)$ have the same secant slope on $f$ as $(1-g,1-r)$ do on $u$.

Since $f$ and $u$ are strictly convex on $[0,1]$, their secant slopes are strictly increasing in each endpoint. In particular, for fixed $g\in[0,1)$,
\begin{equation}
\label{eq:monotone_secant_app}
r\mapsto \frac{r^4-g^4}{r-g} \quad\text{is strictly increasing on } (g,1].
\end{equation}

\subsection{Elimination to one variable}

Using $r^4-g^4=(r-g)(r^3+gr^2+g^2r+g^3)$, the first equation in \eqref{eq:secant_form_app} is equivalent (for $r\ne g$) to the cubic
\begin{equation}
\label{eq:cubic_r_app}
r^3+g r^2+g^2 r+g^3 \;=\; \frac{\mathrm{RE}-g^4}{\overline{\mathrm{VE}}-g}.
\end{equation}
By \eqref{eq:monotone_secant_app}, this has at most one solution $r>g$ for each fixed $g$.

Let $r=R(g)$ denote this unique solution (if it exists) and define
\begin{equation}
\label{eq:E_def_app}
E(g)\;:=\;\frac{(1-R(g))^4-(1-g)^4}{R(g)-g}\;-\;\frac{\mathrm{MA}-(1-g)^4}{\overline{\mathrm{VE}}-g}.
\end{equation}

\paragraph{Lemma (Bijection with $E(g)=0$).}
\emph{For fixed $\overline{\mathrm{VE}}\in(0,1)$, ordered solutions $(r,g)\in\mathcal{D}_{\overline{\mathrm{VE}}}$ of \eqref{eq:secant_form_app} are in one-to-one correspondence with real roots $g\in(0,\overline{\mathrm{VE}})$ of $E(g)=0$ for which $R(g)>\overline{\mathrm{VE}}$. For each such root $g^\star$, the corresponding $r^\star=R(g^\star)$ is unique, and then $\theta^\star=\frac{\overline{\mathrm{VE}}-g^\star}{r^\star-g^\star}$.}

\noindent\textbf{Proof.}
Fix $g\in(0,\overline{\mathrm{VE}})$. The first equality in \eqref{eq:secant_form_app} uniquely determines $r=R(g)>g$ by \eqref{eq:cubic_r_app}; substituting into the second gives $E(g)=0$. Conversely, if $E(g)=0$ and $R(g)>\overline{\mathrm{VE}}$, then $(\theta,r,g)=(\frac{\overline{\mathrm{VE}}-g}{R(g)-g},R(g),g)$ solves \eqref{eq:mixture_system_app}. \hfill$\square$

\paragraph{Remark (Why $R(g)>\overline{\mathrm{VE}}$).}
Since $x\mapsto x^4$ is convex on $[0,1]$, Jensen’s inequality gives $\mathrm{RE}=\theta r^4+(1-\theta)g^4 \ge (\theta r+(1-\theta)g)^4=\overline{\mathrm{VE}}^4$, with strict inequality in the ordered, nondegenerate case $r\ne g$. Hence
\[
\frac{\mathrm{RE}-g^4}{\overline{\mathrm{VE}}-g}\;>\;\frac{\overline{\mathrm{VE}}^4-g^4}{\overline{\mathrm{VE}}-g}.
\]
By strict monotonicity in \eqref{eq:monotone_secant_app}, the unique $r$ satisfying the first secant identity must satisfy $r=R(g)>\overline{\mathrm{VE}}$.

\subsection{The cubic in $g$ and its discriminant}

Clearing denominators in \eqref{eq:secant_form_app} yields a cubic polynomial
\begin{equation}
\label{eq:cubic_g_app}
P_{\mathrm{RE},\overline{\mathrm{VE}},\mathrm{MA}}(g)\;=\;0,
\end{equation}
whose coefficients depend algebraically on $(\mathrm{RE},\overline{\mathrm{VE}},\mathrm{MA})$.  
\emph{Degree justification.} Using \eqref{eq:cubic_r_app}, the first secant identity expresses $\frac{r^4-g^4}{r-g}$ as $r^3+gr^2+g^2r+g^3$, which is linear in the unknown slope $\frac{\mathrm{RE}-g^4}{\overline{\mathrm{VE}}-g}$; substituting this $r=R(g)$ into the second identity and clearing denominators cancels the factor $(r-g)$ and leaves a polynomial of degree at most $3$ in $g$. (Explicit coefficients are lengthy and omitted for brevity.)

By the lemma above, \emph{ordered solutions} are in bijection with \emph{real roots of $P_{\mathrm{RE},\overline{\mathrm{VE}},\mathrm{MA}}(g)$ in $(0,\overline{\mathrm{VE}})$}.

Let $\Delta(P)$ denote the discriminant of a cubic $P(g)=ag^3+bg^2+cg+d$:
\[
\Delta(P)=18abcd-4b^3d+b^2c^2-4ac^3-27a^2d^2.
\]
This determines the real root structure:
\[
\Delta<0\ \Rightarrow\ \text{one real root},\qquad
\Delta>0\ \Rightarrow\ \text{three real roots},\qquad
\Delta=0\ \Rightarrow\ \text{a multiple real root}.
\]

\paragraph{Theorem (Uniqueness certificate).}
\emph{Fix $\overline{\mathrm{VE}}\in(0,1)$ and $(\mathrm{RE},\mathrm{MA})$. Let $P_{\mathrm{RE},\overline{\mathrm{VE}},\mathrm{MA}}$ be as in \eqref{eq:cubic_g_app}.  
If $\Delta(P_{\mathrm{RE},\overline{\mathrm{VE}},\mathrm{MA}})<0$, then there is at most one ordered solution $(r,g)\in\mathcal{D}_{\overline{\mathrm{VE}}}$.  
If, in addition, $P_{\mathrm{RE},\overline{\mathrm{VE}},\mathrm{MA}}$ has a real root $g^\star\in(0,\overline{\mathrm{VE}})$, then
\[
r^\star=R(g^\star),\qquad \theta^\star=\frac{\overline{\mathrm{VE}}-g^\star}{r^\star-g^\star}
\]
gives the unique solution $(\theta^\star,r^\star,g^\star)$ of \eqref{eq:mixture_system_app} up to symmetry.}

\noindent\textbf{Proof.}
Ordered solutions correspond to real roots of $P_{\mathrm{RE},\overline{\mathrm{VE}},\mathrm{MA}}(g)$ in $(0,\overline{\mathrm{VE}})$.  
If $\Delta<0$ then $P$ has a single real root on $\mathbb{R}$, hence at most one in $(0,\overline{\mathrm{VE}})$.  
If such a root exists, the corresponding $(r,g)$ and $\theta$ are uniquely recovered via $R(g)$ and \eqref{eq:ordered_domain_app}. \hfill$\square$

\subsection{Generic uniqueness and the discriminant locus}

Let $\mathcal{R}_{\overline{\mathrm{VE}}}$ be the image of $\mathcal{D}_{\overline{\mathrm{VE}}}$ under the map $(r,g)\mapsto (\mathrm{RE},\mathrm{MA})$ defined by \eqref{eq:secant_form_app}.  
The equation $\Delta(P_{\mathrm{RE},\overline{\mathrm{VE}},\mathrm{MA}})=0$ defines a real algebraic curve $\Sigma_{\overline{\mathrm{VE}}}\subset\mathcal{R}_{\overline{\mathrm{VE}}}$ (the \emph{discriminant locus}).

\paragraph{Theorem (Generic uniqueness).}
\emph{For fixed $\overline{\mathrm{VE}}\in(0,1)$:
\begin{itemize}
\item If $(\mathrm{RE},\mathrm{MA})\in\mathcal{R}_{\overline{\mathrm{VE}}}\setminus\Sigma_{\overline{\mathrm{VE}}}$, then $\Delta<0$ and \eqref{eq:mixture_system_app} has a unique solution $(\theta,r,g)$ up to symmetry.
\item If $(\mathrm{RE},\mathrm{MA})\in\Sigma_{\overline{\mathrm{VE}}}$, then either a multiple solution occurs or three distinct solutions exist.
\end{itemize}
In particular, $\Sigma_{\overline{\mathrm{VE}}}$ has measure zero, so for almost all valid $(\mathrm{RE},\overline{\mathrm{VE}},\mathrm{MA})$ the parameters $(\theta,r,g)$ are uniquely identifiable up to symmetry.}
\medskip

\noindent\textbf{Proof (sketch).}
Off $\Sigma_{\overline{\mathrm{VE}}}$ the simple-root condition $(\partial P/\partial g)\neq 0$ holds generically; by continuity (implicit function theorem), the number of real roots is locally constant and equals $1$, yielding a single ordered solution. On $\Sigma_{\overline{\mathrm{VE}}}$ the discriminant changes sign, creating a multiple or triple real root. \hfill$\square$

\paragraph{Existence note.}
For statistics induced by any nondegenerate mixture in the ordered domain ($r>\overline{\mathrm{VE}}>g$), continuity of the forward map $(r,g)\mapsto(\mathrm{RE},\mathrm{MA})$ and the intermediate value principle ensure that $P_{\mathrm{RE},\overline{\mathrm{VE}},\mathrm{MA}}(g)$ attains a real root in $(0,\overline{\mathrm{VE}})$. Empirically, all rows in our dataset satisfy this condition.

\subsection{Summary}

The system \eqref{eq:mixture_system_app} admits at most three ordered solutions (six with symmetry).  
However, \emph{generically} $\Delta<0$, so there is exactly one ordered solution (and thus one $(\theta,r,g)$ up to symmetry).  
Empirically, our dataset lies in this generic region, which explains why the solver returns either zero or one solution per row.

\section{Additional Evaluation Results}\label{app:additional_eval}
\subsection{Model Scaling Trends}

\begin{figure}[t]
\begin{center}    
\includegraphics[width=0.8\linewidth]{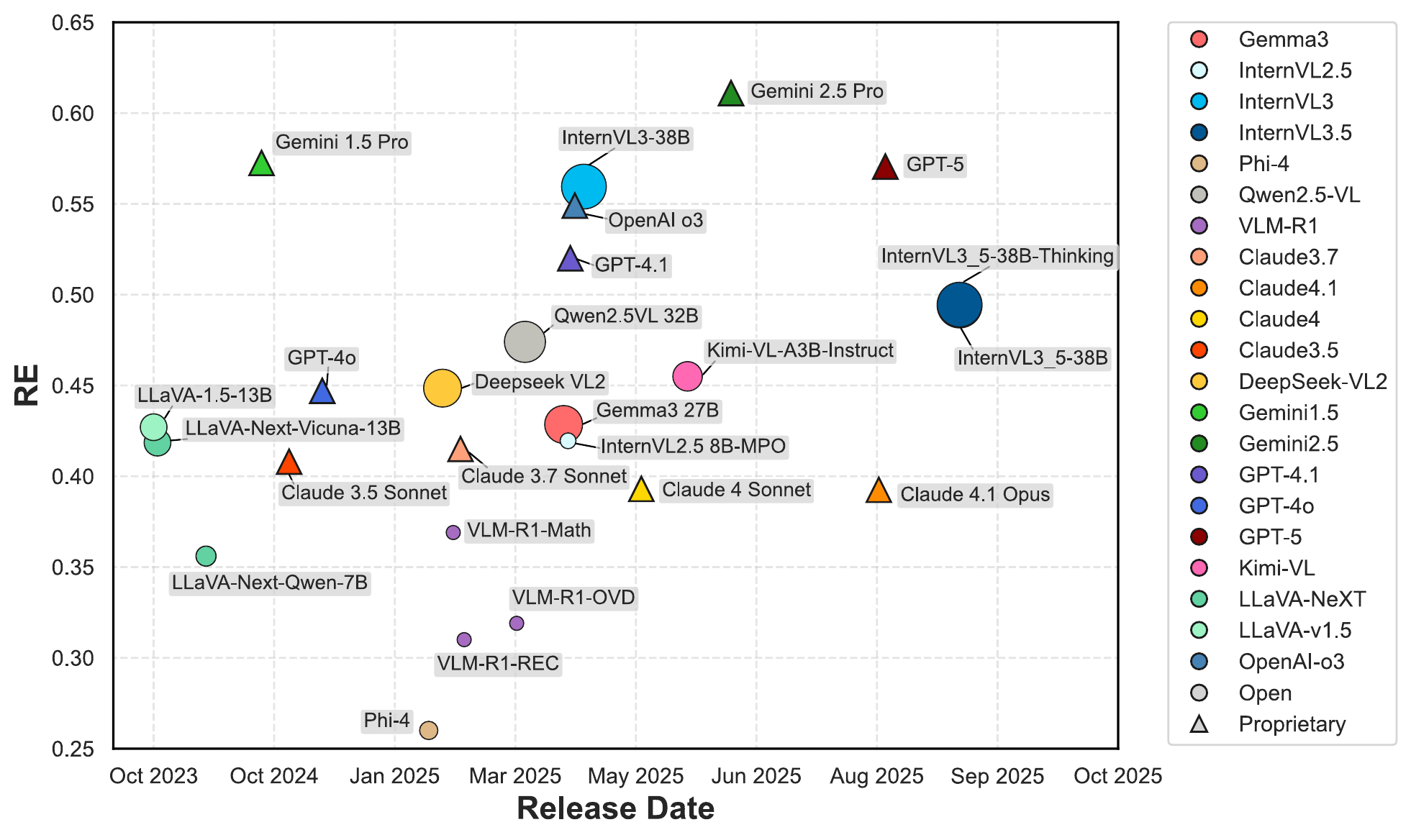}
\end{center}
\caption{Performance (RE) of models versus their release date. Circles denote open-source models, with marker size indicating model scale. Triangles denote proprietary models. Each point represents the largest evaluated model from a given family.}
\label{fig:size_time_performance}  
\end{figure}

We further analyze model performance trends over time and model size. Figure~\ref{fig:re_performance_opensource} shows the relationship between RE performance and model size for various open-source models. To examine temporal trends, Figure~\ref{fig:size_time_performance} plots model performance against their release dates, where open-source models are shown as dots (with marker size indicating model scale) and proprietary models are shown as triangles. Overall, performance tends to rise with newer releases, particularly among proprietary models such as GPT-5 and Gemini 2.5 Pro. Open-source models also progress over time, though less consistently: for instance, InternVL3.5, released after InternVL3, shows no clear RE improvement despite comparable size. A similar pattern appears in the Claude family, where later models (e.g., Claude 4.1 Opus) underperform earlier Sonnet versions on RE. These patterns indicate that top-down visual understanding is not a prioritized objective in current training regimes; most models appear to focus on mainstream capabilities such as chat, long-context reasoning, or coding, while robustness on top-down views receives little explicit attention. This highlights the underexplored status of top-down images and the importance of benchmarks like \pname that bring this gap into focus.

\subsection{Visual Grounding}
\begin{figure}[t]
\begin{center}    
\includegraphics[width=0.9\linewidth]{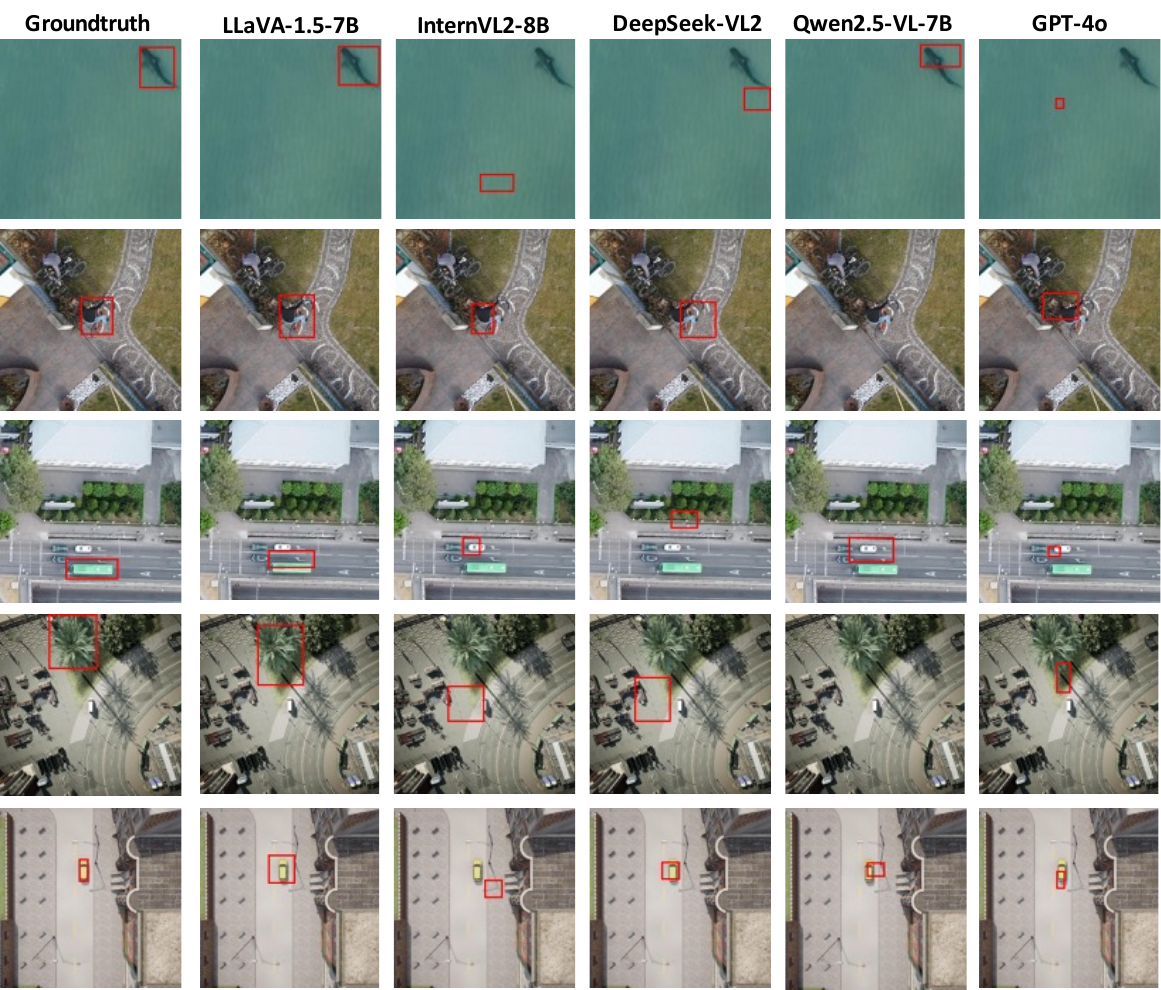}
\end{center}
\caption{Grounding results from various models.}
\label{fig:grounding_results}  
\end{figure}

In \pname, we employ a lenient criteria, centroid containment criterion, for visual grounding evaluation rather than the conventional Intersection over Union (IoU) metric typically used in object detection tasks. The reason is that aerial applications, such as drone navigation scenarios where precise object boundaries are less critical than accurate central positioning as waypoint. Specifically, a prediction is considered successful if the predicted object's centroid falls within the ground truth bounding box, enabling effective target localization for hovering operations. While boundary precision is less relevant in many aerial contexts, we nevertheless present comparative performance analysis using both centroid containment and IoU thresholds in Table~\ref{tab:visual_gnd_method_comp}. Note that value of IoU here is obtained by the calculating the mean in 4 rotations dataset, whereas centroid performance is obtained under RE. We also show some examples of grounding results from some models in Figure~\ref{fig:grounding_results} for reference.

\begin{table}[t]
\centering
\caption{Visual Grounding IoU vs Centroid Containment Comparison.}
\label{tab:visual_gnd_method_comp}
\small
\begin{tabularx}{\textwidth}{l|*{6}{X}}
\hline
\textbf{Metric} 
& \textbf{\begin{tabular}[c]{@{}c@{}}GPT\\4o\end{tabular}} 
& \textbf{\begin{tabular}[c]{@{}c@{}}GPT 4o\\mini\end{tabular}} 
& \textbf{\begin{tabular}[c]{@{}c@{}}Gemini\\1.5 pro\end{tabular}} 
& \textbf{\begin{tabular}[c]{@{}c@{}}Gemini\\1.5 flash\end{tabular}} 
& \textbf{\begin{tabular}[c]{@{}c@{}}Claude 3.5\\sonnet\end{tabular}} 
& \textbf{\begin{tabular}[c]{@{}c@{}}Claude 3.5\\haiku\end{tabular}} \\
\hline
Average IoU & 0.05 & 0.03 & \textbf{0.40} & 0.25 & 0.07 & 0.06 \\
Centroid Performance (\%) & 1.50 & 1.60 & \textbf{36.40} & 24.10 & 2.80 & 1.10 \\
\hline
\end{tabularx}

\vspace{1em}

\begin{tabularx}{\textwidth}{l|*{5}{X}}
\hline
\textbf{Metric} 
& \textbf{\begin{tabular}[c]{@{}c@{}}DeepSeek\\VL2-small\end{tabular}} 
& \textbf{\begin{tabular}[c]{@{}c@{}}DeepSeek\\VL2-tiny\end{tabular}} 
& \textbf{\begin{tabular}[c]{@{}c@{}}LLaVA-Next\\Qwen-7B\end{tabular}} 
& \textbf{\begin{tabular}[c]{@{}c@{}}LLaVA-Next\\Qwen-0.5B\end{tabular}} 
& \textbf{\begin{tabular}[c]{@{}c@{}}LLaVA\\1.5-7B\end{tabular}} \\
\hline
Average IoU & 0.09 & 0.08 & 0.06 & 0.05 & \textbf{0.35} \\
Centroid Performance (\%) & 1.80 & 2.60 & 0.60 & 0.50 & \textbf{36.50} \\
\hline
\end{tabularx}

\vspace{1em}

\begin{tabularx}{\textwidth}{l|*{5}{X}}
\hline
\textbf{Metric} 
& \textbf{\begin{tabular}[c]{@{}c@{}}Qwen2.5\\VL-7B\end{tabular}} 
& \textbf{\begin{tabular}[c]{@{}c@{}}Qwen2.5\\VL-3B\end{tabular}} 
& \textbf{\begin{tabular}[c]{@{}c@{}}InternVL2\\8B\end{tabular}} 
& \textbf{\begin{tabular}[c]{@{}c@{}}InternVL2\\4B\end{tabular}} 
& \textbf{Phi4} \\
\hline
Average IoU & 0.07 & 0.02 & 0.04 & 0.02 & 0.01 \\
Centroid Performance (\%) & 0.60 & 0.00 & 0.50 & 0.00 & 0.00 \\
\hline
\end{tabularx}
\end{table}

\subsection{Comprehensive Dimension-Wise Results}
We have presented only aggregated performance summaries (Figure~\ref{fig:overall_performance}, Table~\ref{tab:rotationalEval} \&~\ref{tab:model_robustness}) in previous sections. 
For completeness, Tables~\ref{tab:detail1}–\ref{tab:detail4} provide the full dimension-wise results of all 60 evaluated models (17 proprietary and 47 open-source) across the 10 evaluation dimensions in \pname. 
Each table contains RE, VE, $\theta$, $r$, $g$, and $A_{\text{adj}}$ for every model, with the best values highlighted in green and the worst in red (separately for open-source and proprietary models). Unlike the other three tables, Table~\ref{tab:detail4} (Visual Grounding) presents only the top 12 models by $A_{\text{adj}}$ in each group. Many models produced near-zero RE and consequently very low $A_{\text{adj}}$ on this task, likely due to the lack of relevant training data, offering little comparative insight. In rare cases, such as for GPT-4o on \textit{Scene Understanding}, a valid solution to the parameter system could not be found.

\begin{table}[t]
\centering
\caption{VLMs in \pname on Scene Understanding, Hallucination Detection, Object Presence.}\label{tab:detail1}
\setlength{\tabcolsep}{5pt}
\renewcommand{\arraystretch}{1.15}
\resizebox{\linewidth}{!}{
\begin{tabular}{l|*{6}{c}|*{6}{c}|*{6}{c}}
\toprule
\textbf{Model} &
\multicolumn{6}{c|}{\textbf{Scene Understanding}} &
\multicolumn{6}{c|}{\textbf{Hallucination Detection}} &
\multicolumn{6}{c}{\textbf{Object Presence}} \\
\cmidrule(lr){2-7}\cmidrule(lr){8-13}\cmidrule(lr){14-19}
& \textbf{RE} & \textbf{VE} & $\boldsymbol{\theta}$ & $\boldsymbol{r}$ & $\boldsymbol{g}$ & \textbf{$A_{\text{adj}}$}
& \textbf{RE} & \textbf{VE} & $\boldsymbol{\theta}$ & $\boldsymbol{r}$ & $\boldsymbol{g}$ & \textbf{$A_{\text{adj}}$}
& \textbf{RE} & \textbf{VE} & $\boldsymbol{\theta}$ & $\boldsymbol{r}$ & $\boldsymbol{g}$ & \textbf{$A_{\text{adj}}$} \\
\midrule
\multicolumn{19}{c}{\textit{Proprietary VLMs}} \\
\midrule
Claude 3.5 Haiku & \worst{0.740} & \worst{0.864} & \worst{0.853} & 0.965 & 0.278 & \worst{0.823} & \best{0.835} & \best{0.901} & 0.884 & \best{0.986} & 0.260 & \best{0.871} & 0.390 & 0.601 & 0.627 & 0.888 & 0.119 & 0.557 \\
Claude 3.5 Sonnet & 0.775 & 0.899 & 0.896 & 0.964 & 0.338 & 0.863 & 0.635 & 0.828 & 0.855 & 0.928 & 0.234 & 0.794 & 0.430 & 0.650 & 0.640 & 0.905 & 0.197 & 0.579 \\
Claude 3.7 Sonnet & 0.780 & 0.892 & 0.861 & 0.975 & 0.384 & 0.839 & 0.745 & 0.881 & \best{0.907} & 0.952 & 0.189 & 0.864 & \worst{0.325} & 0.537 & 0.530 & \worst{0.885} & 0.146 & 0.469 \\
Claude 4 Sonnet & 0.745 & 0.865 & 0.885 & \worst{0.958} & 0.150 & 0.848 & 0.600 & 0.796 & 0.772 & 0.938 & 0.315 & 0.724 & 0.330 & \worst{0.534} & \worst{0.525} & 0.890 & 0.140 & \worst{0.467} \\
Claude 4.1 Opus & 0.800 & 0.899 & 0.896 & 0.972 & 0.266 & 0.871 & 0.515 & 0.743 & 0.742 & 0.912 & 0.254 & 0.677 & 0.340 & 0.550 & 0.526 & 0.896 & 0.165 & 0.472 \\
GPT 4o-mini & 0.870 & 0.934 & 0.940 & 0.981 & 0.197 & 0.922 & 0.745 & 0.875 & 0.846 & 0.968 & 0.365 & 0.819 & 0.465 & 0.635 & 0.642 & 0.923 & 0.120 & 0.592 \\
GPT-4o & 0.930 & 0.961 & - & - & - & - & 0.575 & 0.761 & 0.761 & 0.932 & 0.216 & 0.710 & 0.645 & 0.815 & 0.760 & 0.958 & 0.361 & 0.728 \\
GPT-4.1 Nano & 0.875 & 0.932 & 0.932 & 0.984 & 0.221 & 0.918 & 0.485 & 0.700 & 0.659 & 0.925 & 0.264 & 0.610 & 0.735 & 0.853 & 0.849 & 0.964 & 0.223 & 0.819 \\
GPT-4.1 & 0.915 & 0.961 & 0.943 & 0.992 & 0.456 & 0.935 & 0.405 & 0.629 & 0.642 & 0.891 & 0.158 & 0.572 & 0.725 & 0.855 & 0.848 & 0.961 & 0.263 & 0.815 \\
OpenAI o3 & 0.920 & 0.965 & 0.956 & 0.990 & 0.419 & 0.947 & 0.560 & 0.756 & 0.760 & 0.926 & 0.217 & 0.704 & 0.565 & 0.730 & 0.724 & 0.940 & 0.180 & 0.680 \\
GPT-5 mini & 0.920 & 0.949 & 0.951 & 0.992 & 0.116 & 0.943 & \worst{0.185} & \worst{0.388} & \worst{0.431} & \worst{0.809} & 0.068 & \worst{0.349} & \best{0.865} & 0.919 & 0.906 & \best{0.988} & 0.249 & 0.895 \\
GPT-5 & 0.930 & \best{0.971} & \best{0.973} & 0.989 & 0.344 & 0.962 & 0.550 & 0.730 & 0.707 & 0.939 & 0.226 & 0.664 & 0.615 & 0.754 & 0.731 & 0.958 & 0.201 & 0.700 \\
Gemini 1.5 Flash & 0.905 & 0.948 & 0.953 & 0.987 & 0.147 & 0.941 & 0.540 & 0.703 & 0.708 & 0.934 & 0.139 & 0.662 & 0.720 & 0.815 & 0.811 & 0.971 & 0.147 & 0.787 \\
Gemini 1.5 Pro & 0.920 & 0.953 & 0.956 & 0.991 & 0.134 & 0.947 & 0.525 & 0.714 & 0.719 & 0.924 & 0.175 & 0.664 & 0.810 & 0.886 & 0.882 & 0.979 & 0.193 & 0.864 \\
Gemini 2.5 Flash-Lite & 0.905 & 0.946 & 0.931 & \best{0.993} & 0.316 & 0.925 & 0.655 & 0.820 & 0.835 & 0.941 & 0.208 & 0.786 & 0.745 & 0.843 & 0.839 & 0.971 & 0.173 & 0.815 \\
Gemini 2.5 Flash & 0.920 & 0.956 & 0.953 & 0.991 & 0.249 & 0.945 & 0.590 & 0.784 & 0.729 & 0.947 & 0.345 & 0.690 & 0.770 & 0.866 & 0.875 & 0.969 & 0.151 & 0.847 \\
Gemini 2.5 Pro & \best{0.940} & 0.970 & 0.971 & 0.992 & 0.232 & \best{0.963} & 0.595 & 0.786 & 0.823 & 0.922 & 0.156 & 0.759 & 0.860 & \best{0.930} & \best{0.928} & 0.981 & 0.275 & \best{0.910} \\
\midrule
\multicolumn{19}{c}{\textit{Open Source VLMs}} \\
\midrule
Gemma3 4B & 0.795 & 0.897 & 0.919 & 0.964 & 0.136 & 0.887 & 0.175 & 0.372 & 0.392 & 0.818 & 0.086 & 0.320 & 0.825 & 0.922 & 0.934 & 0.969 & 0.257 & 0.906 \\
Gemma3 12B & 0.780 & 0.896 & 0.907 & 0.963 & 0.246 & 0.873 & 0.255 & 0.477 & 0.477 & 0.855 & 0.134 & 0.407 & 0.805 & 0.894 & 0.894 & 0.974 & 0.216 & 0.871 \\
Gemma3 27B & 0.860 & 0.924 & 0.904 & 0.987 & 0.324 & 0.893 & 0.230 & 0.416 & 0.420 & 0.860 & 0.094 & 0.362 & \best{0.890} & \best{0.949} & \best{0.954} & \best{0.983} & 0.245 & \best{0.937} \\
Deepseek VL2-Tiny & 0.870 & 0.931 & 0.932 & 0.983 & 0.220 & 0.916 & 0.250 & 0.374 & 0.389 & 0.896 & 0.042 & 0.348 & 0.335 & 0.546 & 0.546 & 0.885 & 0.138 & 0.483 \\
Deepseek VL2-Small & 0.885 & 0.932 & 0.913 & \best{0.992} & 0.307 & 0.906 & 0.555 & 0.724 & 0.750 & 0.927 & 0.113 & 0.696 & 0.645 & 0.761 & 0.764 & 0.958 & 0.122 & 0.732 \\
Deepseek VL2 & 0.840 & 0.925 & 0.929 & 0.975 & 0.272 & 0.906 & 0.560 & 0.755 & 0.780 & 0.921 & 0.169 & 0.718 & 0.695 & 0.771 & 0.772 & 0.974 & 0.085 & 0.752 \\
InternVL2.5 4B-MPO & 0.815 & 0.900 & 0.905 & 0.974 & 0.195 & 0.881 & 0.610 & 0.767 & 0.756 & 0.948 & 0.210 & 0.716 & 0.485 & 0.624 & 0.593 & 0.951 & 0.148 & 0.564 \\
InternVL2.5 8B-MPO & 0.810 & 0.881 & 0.848 & 0.988 & 0.284 & 0.838 & 0.625 & 0.785 & 0.809 & 0.937 & 0.139 & 0.758 & 0.415 & 0.594 & 0.573 & 0.922 & 0.153 & 0.528 \\
InternVL3-1B & 0.755 & 0.869 & 0.893 & 0.959 & 0.118 & 0.856 & 0.405 & 0.532 & 0.514 & 0.942 & 0.099 & 0.484 & 0.450 & 0.600 & 0.602 & 0.930 & 0.101 & 0.560 \\
InternVL3-2B & 0.855 & 0.922 & 0.917 & 0.983 & 0.259 & 0.901 & 0.365 & 0.519 & 0.524 & 0.914 & 0.084 & 0.479 & 0.615 & 0.749 & 0.743 & 0.954 & 0.157 & 0.708 \\
InternVL3-8B & 0.880 & 0.924 & 0.932 & 0.986 & 0.074 & 0.919 & 0.405 & 0.546 & 0.554 & 0.925 & 0.076 & 0.512 & 0.595 & 0.759 & 0.777 & 0.935 & 0.144 & 0.727 \\
InternVL3-9B & 0.830 & 0.916 & 0.927 & 0.973 & 0.199 & 0.902 & 0.415 & 0.613 & 0.624 & 0.903 & 0.131 & 0.563 & 0.485 & 0.656 & 0.646 & 0.931 & 0.156 & 0.601 \\
InternVL3-14B & 0.850 & 0.909 & 0.903 & 0.985 & 0.198 & 0.890 & 0.395 & 0.556 & 0.539 & 0.925 & 0.125 & 0.498 & 0.565 & 0.698 & 0.679 & 0.955 & 0.154 & 0.648 \\
InternVL3-38B & \best{0.950} & \best{0.976} & \best{0.982} & 0.992 & 0.140 & \best{0.974} & 0.520 & 0.654 & 0.620 & 0.957 & 0.159 & 0.593 & 0.645 & 0.770 & 0.782 & 0.953 & 0.112 & 0.746 \\
InternVL3.5-1B & 0.705 & 0.821 & 0.829 & 0.960 & 0.149 & 0.796 & 0.220 & 0.396 & 0.402 & 0.860 & 0.085 & 0.346 & 0.645 & 0.811 & 0.794 & 0.949 & 0.282 & 0.753 \\
InternVL3.5-2B & 0.750 & 0.845 & 0.848 & 0.970 & 0.148 & 0.823 & 0.140 & \worst{0.273} & \worst{0.286} & 0.837 & 0.047 & \worst{0.239} & 0.780 & 0.879 & 0.896 & 0.966 & 0.129 & 0.865 \\
InternVL3.5-4B & 0.690 & 0.826 & 0.822 & 0.957 & 0.223 & 0.787 & 0.245 & 0.364 & 0.371 & 0.901 & 0.046 & 0.335 & 0.765 & 0.877 & 0.893 & 0.962 & 0.174 & 0.859 \\
InternVL3.5-8B & 0.700 & 0.834 & 0.819 & 0.961 & 0.258 & 0.787 & \worst{0.130} & 0.275 & 0.307 & 0.807 & 0.040 & 0.247 & 0.720 & 0.839 & 0.791 & 0.976 & 0.319 & 0.772 \\
InternVL3.5-14B & 0.720 & 0.836 & 0.841 & 0.962 & 0.171 & 0.809 & 0.140 & 0.304 & 0.303 & 0.825 & 0.078 & 0.250 & 0.815 & 0.909 & 0.924 & 0.969 & 0.177 & 0.895 \\
InternVL3.5-38B & 0.855 & 0.921 & 0.938 & 0.977 & 0.077 & 0.916 & 0.380 & 0.569 & 0.568 & 0.904 & 0.128 & 0.513 & 0.730 & 0.866 & 0.860 & 0.959 & 0.293 & 0.825 \\
InternVL3.5-1B-Thk & 0.705 & 0.834 & 0.846 & 0.955 & 0.164 & 0.809 & 0.245 & 0.421 & 0.413 & 0.877 & 0.100 & 0.363 & 0.630 & 0.818 & 0.809 & 0.939 & 0.303 & 0.760 \\
InternVL3.5-2B-Thk & 0.720 & 0.820 & 0.807 & 0.972 & 0.185 & 0.784 & 0.230 & 0.502 & 0.559 & \worst{0.801} & 0.125 & 0.448 & 0.545 & 0.767 & 0.808 & 0.906 & 0.185 & 0.732 \\
InternVL3.5-4B-Thk & 0.695 & 0.830 & 0.810 & 0.962 & 0.266 & 0.779 & 0.350 & 0.495 & 0.477 & 0.925 & 0.102 & 0.442 & 0.695 & 0.853 & 0.827 & 0.956 & 0.354 & 0.791 \\
InternVL3.5-8B-Thk & 0.700 & 0.839 & 0.826 & 0.959 & 0.268 & 0.792 & 0.240 & 0.436 & 0.459 & 0.850 & 0.084 & 0.391 & 0.715 & 0.834 & 0.814 & 0.968 & 0.246 & 0.788 \\
InternVL3.5-14B-Thk & 0.730 & 0.860 & 0.860 & 0.960 & 0.247 & 0.826 & 0.200 & 0.424 & 0.476 & 0.805 & 0.078 & 0.383 & 0.780 & 0.889 & 0.865 & 0.974 & 0.345 & 0.842 \\
InternVL3.5-38B-Thk & 0.880 & 0.930 & 0.934 & 0.985 & 0.146 & 0.920 & 0.435 & 0.629 & 0.616 & 0.916 & 0.167 & 0.565 & 0.695 & 0.853 & 0.885 & 0.941 & 0.168 & 0.833 \\
VLM-R1-OVD & 0.615 & 0.786 & \worst{0.778} & 0.943 & 0.237 & 0.734 & 0.370 & 0.593 & 0.583 & 0.892 & 0.173 & 0.520 & 0.440 & 0.636 & 0.669 & 0.901 & 0.102 & 0.602 \\
VLM-R1-Math & 0.645 & 0.799 & 0.791 & 0.950 & 0.226 & 0.752 & 0.505 & 0.671 & 0.677 & 0.929 & 0.130 & 0.629 & 0.480 & 0.667 & 0.684 & 0.915 & 0.131 & 0.626 \\
VLM-R1-REC & \worst{0.580} & \worst{0.777} & 0.785 & \worst{0.927} & 0.231 & \worst{0.728} & 0.530 & 0.730 & 0.752 & 0.916 & 0.166 & 0.689 & 0.305 & 0.535 & 0.533 & 0.870 & 0.154 & 0.463 \\
Kimi-VL-A3B-Thk & 0.735 & 0.863 & 0.840 & 0.967 & 0.316 & 0.812 & 0.355 & 0.611 & 0.571 & 0.887 & 0.244 & 0.506 & 0.375 & 0.601 & 0.607 & 0.886 & 0.161 & 0.538 \\
Kimi-VL-A3B-Instruct & 0.850 & 0.926 & 0.908 & 0.983 & 0.365 & 0.893 & 0.625 & 0.761 & 0.764 & 0.951 & 0.147 & 0.727 & 0.630 & 0.746 & 0.745 & 0.959 & 0.124 & 0.715 \\
Kimi-VL-A3B-Thk-2506 & 0.875 & 0.934 & 0.936 & 0.983 & 0.209 & 0.920 & \best{0.725} & \best{0.858} & 0.812 & \best{0.971} & 0.368 & 0.788 & 0.420 & 0.591 & 0.614 & 0.910 & 0.086 & 0.558 \\
LLaVA-Interleave-Qwen-0.5B & 0.710 & 0.797 & 0.792 & 0.973 & 0.128 & 0.771 & 0.205 & 0.334 & 0.345 & 0.878 & 0.047 & 0.303 & 0.460 & 0.604 & 0.584 & 0.942 & 0.128 & 0.550 \\
LLaVA-1.5-7B & 0.810 & 0.890 & 0.878 & 0.980 & 0.244 & 0.860 & 0.280 & 0.529 & 0.545 & 0.846 & 0.148 & 0.461 & 0.320 & 0.545 & 0.541 & 0.877 & 0.154 & 0.474 \\
LLaVA-Next-Mistral-7B & 0.820 & 0.896 & 0.889 & 0.980 & 0.225 & 0.871 & 0.585 & 0.704 & 0.710 & 0.953 & 0.095 & 0.676 & 0.755 & 0.836 & 0.842 & 0.973 & 0.109 & 0.819 \\
LLaVA-Next-Vicuna-7B & 0.720 & 0.829 & 0.819 & 0.968 & 0.198 & 0.793 & 0.340 & 0.481 & 0.496 & 0.910 & 0.059 & 0.452 & 0.520 & 0.701 & 0.711 & 0.925 & 0.151 & 0.658 \\
LLaVA-Interleave-Qwen-7B & 0.895 & 0.946 & 0.923 & 0.992 & 0.399 & 0.916 & 0.420 & 0.594 & 0.620 & 0.907 & 0.082 & 0.563 & 0.570 & 0.696 & 0.692 & 0.952 & 0.119 & 0.660 \\
LLaVA-1.5-13B & 0.760 & 0.873 & 0.874 & 0.966 & 0.227 & 0.844 & 0.450 & 0.620 & 0.659 & 0.909 & 0.061 & 0.599 & 0.620 & 0.776 & 0.774 & 0.946 & 0.195 & 0.732 \\
LLaVA-Next-Vicuna-13B & 0.725 & 0.829 & 0.815 & 0.971 & 0.202 & 0.791 & 0.390 & 0.530 & 0.533 & 0.925 & 0.080 & 0.493 & 0.605 & 0.706 & 0.714 & 0.959 & 0.074 & 0.685 \\
Phi-4 & 0.680 & 0.812 & 0.811 & 0.957 & 0.194 & 0.776 & 0.705 & 0.833 & \best{0.849} & 0.955 & 0.147 & \best{0.810} & \worst{0.140} & \worst{0.304} & \worst{0.287} & \worst{0.836} & 0.090 & \worst{0.240} \\
Qwen2.5VL 3B & 0.665 & 0.804 & 0.788 & 0.958 & 0.228 & 0.755 & 0.590 & 0.752 & 0.763 & 0.938 & 0.157 & 0.715 & 0.470 & 0.665 & 0.668 & 0.916 & 0.160 & 0.612 \\
Qwen2.5VL 7B & 0.825 & 0.914 & 0.901 & 0.978 & 0.329 & 0.881 & 0.720 & 0.834 & 0.810 & 0.971 & 0.251 & 0.786 & 0.435 & 0.590 & 0.583 & 0.929 & 0.116 & 0.542 \\
Qwen2.5VL 32B & 0.670 & 0.838 & 0.861 & 0.939 & 0.206 & 0.809 & 0.655 & 0.789 & 0.800 & 0.951 & 0.139 & 0.761 & 0.435 & 0.611 & 0.594 & 0.925 & 0.152 & 0.550 \\
\bottomrule
\end{tabular}}
\end{table}

\begin{table}[t]
\centering
\caption{VLMs in \pname on Object Localization, Attribute Recognition, Object Counting.}\label{tab:detail2}
\setlength{\tabcolsep}{5pt}
\renewcommand{\arraystretch}{1.15}
\resizebox{\linewidth}{!}{
\begin{tabular}{l|*{6}{c}|*{6}{c}|*{6}{c}}
\toprule
\textbf{Model} &
\multicolumn{6}{c|}{\textbf{Object Localization}} &
\multicolumn{6}{c|}{\textbf{Attribute Recognition}} &
\multicolumn{6}{c}{\textbf{Object Counting}} \\
\cmidrule(lr){2-7}\cmidrule(lr){8-13}\cmidrule(lr){14-19}
& \textbf{RE} & \textbf{VE} & $\boldsymbol{\theta}$ & $\boldsymbol{r}$ & $\boldsymbol{g}$ & \textbf{$A_{\text{adj}}$}
& \textbf{RE} & \textbf{VE} & $\boldsymbol{\theta}$ & $\boldsymbol{r}$ & $\boldsymbol{g}$ & \textbf{$A_{\text{adj}}$}
& \textbf{RE} & \textbf{VE} & $\boldsymbol{\theta}$ & $\boldsymbol{r}$ & $\boldsymbol{g}$ & \textbf{$A_{\text{adj}}$} \\
\midrule
\multicolumn{19}{c}{\textit{Proprietary VLMs}} \\
\midrule
Claude 3.5 Haiku & 0.165 & 0.496 & \worst{0.475} & 0.765 & 0.253 & 0.363 & \worst{0.405} & \worst{0.608} & \worst{0.579} & 0.914 & 0.186 & \worst{0.529} & \worst{0.075} & \worst{0.316} & 0.373 & \worst{0.669} & 0.106 & \worst{0.250} \\
Claude 3.5 Sonnet & 0.335 & 0.627 & 0.541 & 0.884 & 0.326 & 0.478 & 0.525 & 0.714 & 0.711 & 0.927 & 0.189 & 0.659 & 0.135 & 0.394 & 0.397 & 0.763 & 0.151 & 0.303 \\
Claude 3.7 Sonnet & 0.340 & 0.583 & 0.484 & 0.914 & 0.272 & 0.442 & 0.480 & 0.669 & 0.613 & 0.940 & 0.239 & 0.577 & 0.115 & 0.354 & 0.388 & 0.738 & 0.111 & 0.286 \\
Claude 4 Sonnet & 0.420 & 0.693 & 0.664 & 0.890 & 0.302 & 0.591 & 0.490 & 0.704 & 0.737 & \worst{0.903} & 0.146 & 0.665 & 0.100 & 0.399 & 0.427 & 0.695 & 0.178 & 0.297 \\
Claude 4.1 Opus & 0.365 & 0.641 & 0.623 & 0.874 & 0.257 & 0.544 & 0.505 & 0.703 & 0.717 & 0.916 & 0.161 & 0.657 & 0.165 & 0.398 & 0.425 & 0.789 & 0.108 & 0.335 \\
GPT 4o-mini & \worst{0.075} & \worst{0.468} & 0.488 & \worst{0.613} & 0.328 & \worst{0.299} & 0.480 & 0.693 & 0.659 & 0.923 & 0.246 & 0.609 & 0.175 & 0.366 & \worst{0.303} & 0.871 & 0.146 & 0.264 \\
GPT-4o & 0.435 & 0.728 & 0.722 & 0.879 & 0.334 & 0.635 & 0.610 & 0.796 & 0.800 & 0.934 & 0.245 & 0.747 & 0.200 & 0.465 & 0.370 & 0.855 & 0.236 & 0.316 \\
GPT-4.1 Nano & 0.570 & 0.811 & 0.874 & 0.899 & 0.207 & 0.785 & 0.700 & 0.839 & 0.865 & 0.949 & 0.137 & 0.820 & 0.185 & 0.453 & 0.411 & 0.818 & 0.197 & 0.336 \\
GPT-4.1 & 0.660 & 0.839 & 0.846 & 0.940 & 0.287 & 0.795 & 0.680 & 0.818 & 0.779 & 0.966 & 0.293 & 0.753 & 0.235 & 0.477 & 0.437 & 0.856 & 0.184 & 0.374 \\
OpenAI o3 & 0.780 & 0.891 & 0.900 & 0.965 & 0.232 & 0.868 & 0.720 & 0.838 & 0.812 & 0.970 & 0.264 & 0.788 & 0.215 & 0.480 & 0.421 & 0.844 & 0.215 & 0.356 \\
GPT-5 mini & 0.575 & 0.826 & 0.831 & 0.910 & 0.414 & 0.756 & 0.700 & 0.821 & 0.827 & 0.959 & 0.163 & 0.793 & 0.215 & 0.484 & 0.449 & 0.831 & 0.201 & 0.373 \\
GPT-5 & 0.770 & 0.887 & 0.886 & \best{0.965} & 0.284 & 0.855 & 0.700 & 0.838 & 0.841 & 0.955 & 0.217 & 0.803 & 0.190 & 0.465 & 0.436 & 0.812 & 0.197 & 0.354 \\
Gemini 1.5 Flash & 0.600 & 0.841 & 0.926 & 0.897 & 0.144 & 0.831 & 0.780 & 0.869 & 0.861 & \best{0.975} & 0.207 & 0.840 & 0.255 & 0.494 & \best{0.506} & 0.842 & 0.136 & \best{0.426} \\
Gemini 1.5 Pro & 0.715 & 0.892 & \best{0.928} & 0.937 & 0.325 & \best{0.869} & 0.740 & 0.860 & 0.851 & 0.966 & 0.259 & 0.821 & \best{0.285} & 0.492 & 0.448 & \best{0.893} & 0.168 & 0.400 \\
Gemini 2.5 Flash-Lite & 0.460 & 0.721 & 0.744 & 0.886 & 0.241 & 0.660 & 0.695 & 0.819 & 0.792 & 0.968 & 0.252 & 0.766 & 0.125 & 0.374 & 0.376 & 0.759 & 0.142 & 0.285 \\
Gemini 2.5 Flash & 0.585 & 0.795 & 0.685 & 0.956 & 0.445 & 0.655 & 0.755 & 0.856 & 0.834 & 0.975 & 0.259 & 0.813 & 0.145 & 0.432 & 0.422 & 0.765 & 0.190 & 0.323 \\
Gemini 2.5 Pro & \best{0.780} & \best{0.901} & 0.900 & 0.964 & 0.331 & 0.868 & \best{0.805} & \best{0.900} & \best{0.913} & 0.969 & 0.177 & \best{0.885} & 0.210 & \best{0.499} & 0.483 & 0.811 & 0.207 & 0.392 \\
\midrule
\multicolumn{19}{c}{\textit{Open Source VLMs}} \\
\midrule
Gemma3 4B & 0.035 & 0.400 & 0.074 & 0.684 & 0.377 & 0.051 & 0.435 & 0.677 & 0.698 & 0.888 & 0.190 & 0.620 & \worst{0.035} & 0.301 & 0.281 & \worst{0.590} & 0.188 & 0.166 \\
Gemma3 12B & 0.280 & 0.593 & 0.453 & 0.880 & 0.355 & 0.399 & \worst{0.290} & 0.666 & \best{0.867} & \best{0.761} & 0.053 & 0.659 & 0.130 & 0.414 & 0.438 & 0.738 & 0.161 & 0.323 \\
Gemma3 27B & 0.440 & 0.689 & 0.646 & 0.907 & 0.290 & 0.586 & 0.590 & 0.770 & 0.757 & 0.939 & 0.242 & 0.711 & 0.125 & 0.365 & 0.363 & 0.766 & 0.137 & 0.278 \\
Deepseek VL2-Tiny & 0.130 & 0.410 & 0.395 & 0.756 & 0.184 & 0.299 & 0.610 & 0.774 & 0.789 & 0.938 & 0.160 & 0.740 & 0.165 & 0.369 & 0.319 & 0.848 & 0.145 & 0.270 \\
Deepseek VL2-Small & 0.375 & 0.705 & 0.697 & 0.853 & 0.364 & 0.595 & 0.725 & 0.831 & 0.835 & 0.965 & 0.153 & 0.806 & 0.235 & 0.395 & 0.361 & 0.898 & 0.110 & 0.325 \\
Deepseek VL2 & 0.365 & 0.723 & 0.820 & 0.816 & 0.297 & 0.669 & 0.680 & 0.830 & 0.857 & 0.944 & 0.150 & 0.809 & 0.235 & 0.398 & 0.358 & 0.900 & 0.117 & 0.322 \\
InternVL2.5 4B-MPO & 0.180 & 0.531 & 0.363 & 0.826 & 0.363 & 0.300 & 0.570 & 0.754 & 0.741 & 0.936 & 0.233 & 0.693 & 0.260 & 0.432 & 0.415 & 0.890 & 0.108 & 0.369 \\
InternVL2.5 8B-MPO & 0.390 & 0.649 & 0.616 & 0.891 & 0.261 & 0.549 & 0.630 & 0.784 & 0.781 & 0.948 & 0.199 & 0.740 & 0.230 & 0.434 & 0.443 & 0.849 & 0.103 & 0.376 \\
InternVL3-1B & 0.110 & 0.459 & 0.441 & 0.702 & 0.267 & 0.309 & 0.500 & 0.699 & 0.696 & 0.921 & 0.192 & 0.640 & 0.300 & 0.427 & 0.446 & 0.906 & 0.043 & 0.404 \\
InternVL3-2B & 0.270 & 0.534 & 0.502 & 0.856 & 0.209 & 0.430 & 0.595 & 0.761 & 0.741 & 0.946 & 0.233 & 0.701 & 0.285 & 0.435 & 0.448 & 0.893 & 0.063 & 0.400 \\
InternVL3-8B & 0.570 & 0.769 & 0.724 & 0.941 & 0.317 & 0.681 & 0.660 & 0.807 & 0.820 & 0.947 & 0.171 & 0.777 & 0.165 & 0.414 & 0.454 & 0.776 & 0.113 & 0.352 \\
InternVL3-9B & 0.640 & 0.815 & 0.767 & 0.954 & 0.356 & 0.732 & 0.630 & 0.794 & 0.789 & 0.945 & 0.228 & 0.746 & 0.315 & 0.465 & 0.454 & 0.913 & 0.093 & 0.414 \\
InternVL3-14B & 0.595 & 0.823 & 0.883 & 0.906 & 0.191 & 0.800 & 0.650 & 0.792 & 0.783 & 0.954 & 0.209 & 0.747 & 0.320 & 0.490 & 0.496 & 0.896 & 0.091 & 0.444 \\
InternVL3-38B & 0.795 & 0.911 & 0.907 & 0.967 & 0.366 & 0.877 & \best{0.800} & \best{0.874} & 0.860 & \best{0.982} & 0.208 & \best{0.845} & 0.340 & 0.475 & 0.455 & 0.930 & 0.095 & 0.423 \\
InternVL3.5-1B & 0.330 & 0.588 & 0.566 & 0.873 & 0.215 & 0.494 & 0.540 & 0.703 & 0.729 & 0.928 & 0.098 & 0.676 & \best{0.360} & 0.482 & 0.479 & \best{0.931} & 0.070 & 0.446 \\
InternVL3.5-2B & 0.410 & 0.679 & 0.567 & 0.918 & 0.366 & 0.520 & 0.525 & 0.730 & 0.715 & 0.925 & 0.240 & 0.662 & 0.315 & 0.474 & 0.474 & 0.903 & 0.087 & 0.428 \\
InternVL3.5-4B & 0.625 & 0.815 & 0.805 & 0.938 & 0.307 & 0.755 & 0.555 & 0.720 & 0.716 & 0.938 & 0.170 & 0.672 & 0.280 & 0.487 & 0.478 & 0.875 & 0.133 & 0.418 \\
InternVL3.5-8B & 0.730 & 0.875 & 0.881 & 0.954 & 0.291 & 0.840 & 0.540 & 0.733 & 0.734 & 0.926 & 0.198 & 0.680 & 0.320 & 0.504 & 0.478 & 0.904 & 0.137 & 0.432 \\
InternVL3.5-14B & 0.710 & 0.874 & 0.899 & 0.943 & 0.262 & 0.847 & 0.515 & 0.709 & 0.697 & 0.927 & 0.207 & 0.646 & 0.300 & 0.499 & 0.523 & 0.870 & 0.092 & 0.455 \\
InternVL3.5-38B & \best{0.820} & \best{0.919} & \best{0.935} & \best{0.968} & 0.211 & \best{0.905} & 0.495 & 0.713 & 0.724 & 0.909 & 0.197 & 0.658 & 0.355 & \best{0.564} & \best{0.533} & 0.903 & 0.176 & \best{0.481} \\
InternVL3.5-1B-Thk & 0.210 & 0.459 & 0.447 & 0.827 & 0.160 & 0.370 & 0.525 & 0.693 & 0.704 & 0.929 & 0.130 & 0.654 & 0.185 & 0.429 & 0.379 & 0.835 & 0.181 & 0.316 \\
InternVL3.5-2B-Thk & 0.245 & 0.532 & 0.405 & 0.878 & 0.297 & 0.356 & 0.435 & 0.679 & 0.711 & 0.884 & 0.174 & 0.628 & 0.185 & 0.444 & 0.450 & 0.801 & 0.152 & 0.360 \\
InternVL3.5-4B-Thk & 0.525 & 0.781 & 0.814 & 0.896 & 0.280 & 0.729 & 0.555 & 0.734 & 0.723 & 0.936 & 0.206 & 0.677 & 0.175 & 0.451 & 0.477 & 0.778 & 0.153 & 0.371 \\
InternVL3.5-8B-Thk & 0.670 & 0.853 & 0.861 & 0.939 & 0.319 & 0.808 & 0.570 & 0.761 & 0.789 & 0.922 & 0.160 & 0.728 & 0.260 & 0.510 & 0.529 & 0.837 & 0.142 & 0.443 \\
InternVL3.5-14B-Thk & 0.650 & 0.853 & 0.805 & 0.944 & 0.474 & 0.760 & 0.570 & 0.748 & 0.760 & 0.931 & 0.168 & 0.707 & 0.325 & 0.539 & 0.513 & 0.892 & 0.167 & 0.458 \\
InternVL3.5-38B-Thk & 0.790 & 0.901 & 0.914 & 0.964 & 0.231 & 0.882 & 0.555 & 0.761 & 0.742 & 0.929 & 0.279 & 0.689 & 0.325 & 0.545 & 0.527 & 0.886 & 0.165 & 0.467 \\
VLM-R1-OVD & 0.445 & 0.731 & 0.698 & 0.891 & 0.363 & 0.622 & 0.525 & 0.738 & 0.742 & 0.917 & 0.221 & 0.681 & 0.160 & 0.407 & 0.416 & 0.787 & 0.137 & 0.328 \\
VLM-R1-Math & 0.495 & 0.772 & 0.668 & 0.920 & 0.476 & 0.614 & 0.585 & 0.764 & 0.759 & 0.937 & 0.218 & 0.711 & 0.145 & 0.367 & 0.347 & 0.804 & 0.136 & 0.279 \\
VLM-R1-REC & 0.330 & 0.641 & 0.561 & 0.871 & 0.347 & 0.489 & 0.455 & 0.711 & 0.752 & 0.882 & 0.193 & 0.663 & 0.120 & 0.354 & 0.298 & 0.796 & 0.166 & 0.237 \\
Kimi-VL-A3B-Thk & 0.160 & 0.455 & 0.330 & 0.830 & 0.271 & 0.274 & 0.555 & 0.749 & 0.749 & 0.928 & 0.216 & 0.694 & 0.060 & 0.368 & 0.423 & 0.612 & 0.188 & 0.259 \\
Kimi-VL-A3B-Instruct & 0.555 & 0.776 & 0.800 & 0.912 & 0.231 & 0.730 & 0.710 & 0.825 & 0.835 & 0.960 & 0.141 & 0.802 & 0.260 & 0.441 & 0.431 & 0.881 & 0.108 & 0.380 \\
Kimi-VL-A3B-Thk-2506 & 0.510 & 0.746 & 0.686 & 0.926 & 0.353 & 0.635 & 0.650 & 0.806 & 0.793 & 0.951 & 0.252 & 0.754 & 0.120 & 0.331 & 0.327 & 0.778 & 0.114 & 0.254 \\
LLaVA-Interleave-Qwen-0.5B & 0.015 & 0.270 & \worst{0.015} & 0.917 & 0.260 & \worst{0.014} & 0.420 & 0.630 & 0.629 & 0.904 & 0.165 & 0.569 & 0.060 & \worst{0.217} & \worst{0.184} & 0.756 & 0.096 & \worst{0.139} \\
LLaVA-1.5-7B & 0.115 & 0.535 & 0.733 & \worst{0.627} & 0.282 & 0.460 & 0.385 & 0.644 & 0.638 & 0.881 & 0.226 & 0.562 & 0.180 & 0.361 & 0.359 & 0.842 & 0.093 & 0.302 \\
LLaVA-Next-Mistral-7B & 0.700 & 0.853 & 0.853 & 0.951 & 0.278 & 0.812 & 0.690 & 0.820 & 0.837 & 0.953 & 0.138 & 0.797 & 0.110 & 0.314 & 0.277 & 0.793 & 0.130 & 0.220 \\
LLaVA-Next-Vicuna-7B & 0.435 & 0.704 & 0.609 & 0.915 & 0.374 & 0.558 & 0.575 & 0.750 & 0.749 & 0.936 & 0.196 & 0.701 & 0.105 & 0.314 & 0.295 & 0.772 & 0.122 & 0.228 \\
LLaVA-Interleave-Qwen-7B & 0.170 & 0.514 & 0.472 & 0.771 & 0.284 & 0.364 & 0.660 & 0.807 & 0.813 & 0.949 & 0.191 & 0.772 & 0.175 & 0.351 & 0.322 & 0.858 & 0.110 & 0.277 \\
LLaVA-1.5-13B & 0.385 & 0.718 & 0.778 & 0.838 & 0.297 & 0.652 & 0.535 & 0.733 & 0.747 & 0.920 & 0.179 & 0.687 & 0.205 & 0.345 & 0.326 & 0.891 & 0.082 & 0.290 \\
LLaVA-Next-Vicuna-13B & 0.470 & 0.741 & 0.706 & 0.901 & 0.358 & 0.636 & 0.660 & 0.789 & 0.801 & 0.953 & 0.128 & 0.763 & 0.085 & 0.278 & 0.212 & 0.795 & 0.138 & 0.169 \\
Phi-4 & \worst{0.010} & \worst{0.244} & - & - & - & - & 0.385 & \worst{0.600} & \worst{0.598} & 0.896 & 0.160 & \worst{0.536} & 0.050 & 0.224 & 0.225 & 0.686 & 0.089 & 0.154 \\
Qwen2.5VL 3B & 0.470 & 0.726 & 0.709 & 0.901 & 0.300 & 0.639 & 0.595 & 0.762 & 0.762 & 0.940 & 0.195 & 0.716 & 0.185 & 0.401 & 0.391 & 0.829 & 0.126 & 0.324 \\
Qwen2.5VL 7B & 0.715 & 0.863 & 0.862 & 0.954 & 0.290 & 0.823 & 0.665 & 0.825 & 0.780 & 0.960 & 0.347 & 0.749 & 0.125 & 0.326 & 0.311 & 0.796 & 0.114 & 0.248 \\
Qwen2.5VL 32B & 0.600 & 0.824 & 0.757 & 0.939 & 0.464 & 0.711 & 0.630 & 0.814 & 0.840 & 0.931 & 0.202 & 0.781 & 0.115 & 0.321 & 0.282 & 0.799 & 0.134 & 0.225 \\
\bottomrule
\end{tabular}}
\end{table}

\begin{table}[t]
\centering
\caption{VLMs in \pname on Attribute Comparison, Dynamic Temporal, Spatial Relationship.}\label{tab:detail3}
\setlength{\tabcolsep}{5pt}
\renewcommand{\arraystretch}{1.15}
\resizebox{\linewidth}{!}{
\begin{tabular}{l|*{6}{c}|*{6}{c}|*{6}{c}}
\toprule
\textbf{Model} &
\multicolumn{6}{c|}{\textbf{Attribute Comparison}} &
\multicolumn{6}{c|}{\textbf{Dynamic Temporal}} &
\multicolumn{6}{c}{\textbf{Spatial Relationship}} \\
\cmidrule(lr){2-7}\cmidrule(lr){8-13}\cmidrule(lr){14-19}
& \textbf{RE} & \textbf{VE} & $\boldsymbol{\theta}$ & $\boldsymbol{r}$ & $\boldsymbol{g}$ & \textbf{$A_{\text{adj}}$}
& \textbf{RE} & \textbf{VE} & $\boldsymbol{\theta}$ & $\boldsymbol{r}$ & $\boldsymbol{g}$ & \textbf{$A_{\text{adj}}$}
& \textbf{RE} & \textbf{VE} & $\boldsymbol{\theta}$ & $\boldsymbol{r}$ & $\boldsymbol{g}$ & \textbf{$A_{\text{adj}}$} \\
\midrule
\multicolumn{19}{c}{\textit{Proprietary VLMs}} \\
\midrule
Claude 3.5 Haiku & \best{0.615} & 0.670 & 0.669 & \best{0.979} & 0.045 & 0.655 & 0.200 & \worst{0.375} & 0.333 & 0.880 & 0.123 & \worst{0.293} & 0.190 & 0.501 & 0.576 & 0.758 & 0.153 & 0.436 \\
Claude 3.5 Sonnet & 0.510 & 0.669 & 0.676 & 0.932 & 0.119 & 0.630 & 0.295 & 0.610 & 0.501 & 0.871 & 0.349 & 0.436 & 0.410 & 0.715 & \worst{0.539} & 0.920 & 0.475 & 0.496 \\
Claude 3.7 Sonnet & 0.610 & 0.693 & 0.696 & 0.967 & 0.062 & 0.674 & 0.245 & 0.619 & 0.505 & 0.822 & 0.411 & 0.415 & 0.475 & 0.723 & 0.700 & 0.907 & 0.293 & 0.635 \\
Claude 4 Sonnet & 0.545 & 0.679 & 0.708 & 0.937 & 0.053 & 0.663 & 0.200 & 0.519 & 0.468 & 0.806 & 0.266 & 0.377 & 0.480 & 0.759 & 0.782 & 0.884 & 0.308 & 0.692 \\
Claude 4.1 Opus & 0.530 & 0.679 & 0.687 & 0.937 & 0.112 & 0.644 & 0.210 & 0.530 & 0.471 & 0.814 & 0.277 & 0.383 & 0.485 & 0.776 & 0.867 & 0.865 & 0.199 & 0.750 \\
GPT 4o-mini & 0.420 & \worst{0.630} & 0.660 & 0.893 & 0.119 & 0.590 & \worst{0.145} & 0.432 & 0.383 & 0.783 & 0.215 & 0.300 & \worst{0.025} & \worst{0.393} & \best{0.983} & \worst{0.399} & 0.004 & \worst{0.392} \\
GPT-4o & 0.420 & 0.647 & \worst{0.634} & 0.902 & 0.207 & \worst{0.572} & 0.225 & 0.529 & 0.444 & 0.840 & 0.280 & 0.373 & 0.415 & 0.720 & 0.695 & 0.876 & 0.364 & 0.609 \\
GPT-4.1 Nano & 0.600 & 0.703 & 0.708 & 0.959 & 0.079 & 0.679 & 0.310 & 0.627 & 0.456 & 0.898 & 0.401 & 0.409 & 0.665 & 0.860 & 0.863 & 0.936 & 0.382 & 0.808 \\
GPT-4.1 & 0.525 & 0.677 & 0.683 & 0.936 & 0.120 & 0.639 & 0.240 & 0.536 & 0.483 & 0.838 & 0.254 & 0.405 & 0.600 & 0.815 & 0.691 & 0.958 & 0.495 & 0.662 \\
OpenAI o3 & 0.550 & 0.714 & 0.694 & 0.943 & 0.192 & 0.655 & 0.280 & 0.605 & 0.409 & 0.897 & 0.403 & 0.367 & 0.770 & 0.909 & 0.903 & 0.960 & 0.434 & 0.867 \\
GPT-5 mini & 0.540 & 0.705 & 0.710 & 0.934 & 0.145 & 0.663 & 0.210 & 0.481 & 0.413 & 0.843 & 0.227 & 0.348 & \best{0.855} & \best{0.934} & 0.951 & \best{0.974} & 0.153 & \best{0.926} \\
GPT-5 & 0.580 & 0.733 & 0.731 & 0.944 & 0.158 & 0.690 & \best{0.340} & \best{0.679} & \worst{0.324} & \best{0.966} & 0.541 & 0.313 & 0.805 & 0.922 & 0.958 & 0.957 & 0.119 & 0.918 \\
Gemini 1.5 Flash & 0.475 & 0.649 & 0.671 & 0.917 & 0.101 & 0.616 & 0.190 & 0.489 & 0.440 & 0.809 & 0.237 & 0.356 & 0.635 & 0.829 & 0.819 & 0.938 & 0.337 & 0.768 \\
Gemini 1.5 Pro & 0.565 & 0.704 & 0.696 & 0.949 & 0.142 & 0.661 & 0.185 & 0.469 & 0.485 & 0.785 & 0.170 & 0.381 & 0.620 & 0.845 & 0.857 & 0.921 & 0.388 & 0.790 \\
Gemini 2.5 Flash-Lite & \worst{0.410} & 0.639 & 0.656 & \worst{0.889} & 0.162 & 0.583 & 0.170 & 0.505 & 0.455 & \worst{0.778} & 0.277 & 0.354 & 0.620 & 0.846 & 0.865 & 0.919 & 0.380 & 0.795 \\
Gemini 2.5 Flash & 0.455 & 0.691 & 0.663 & 0.909 & 0.262 & 0.603 & 0.225 & 0.525 & 0.515 & 0.812 & 0.220 & 0.418 & 0.820 & 0.921 & 0.943 & 0.966 & 0.186 & 0.911 \\
Gemini 2.5 Pro & 0.575 & \best{0.744} & \best{0.743} & 0.938 & 0.182 & \best{0.697} & 0.240 & 0.535 & \best{0.542} & 0.815 & 0.203 & \best{0.442} & 0.825 & 0.925 & 0.943 & 0.967 & 0.231 & 0.912 \\
\midrule
\multicolumn{19}{c}{\textit{Open Source VLMs}} \\
\midrule
Gemma3 4B & 0.260 & 0.550 & 0.523 & 0.838 & 0.233 & 0.439 & 0.140 & 0.429 & 0.463 & 0.741 & 0.160 & 0.343 & 0.020 & 0.331 & 0.567 & 0.429 & 0.203 & 0.243 \\
Gemma3 12B & 0.450 & 0.634 & 0.610 & 0.927 & 0.176 & 0.565 & 0.165 & 0.415 & 0.472 & 0.769 & 0.098 & 0.363 & 0.135 & 0.550 & 0.816 & 0.638 & 0.162 & 0.520 \\
Gemma3 27B & 0.490 & 0.676 & 0.670 & 0.925 & 0.172 & 0.619 & 0.145 & 0.386 & 0.422 & 0.766 & 0.109 & 0.323 & 0.495 & 0.744 & 0.673 & 0.923 & 0.375 & 0.621 \\
Deepseek VL2-Tiny & 0.330 & 0.603 & 0.607 & 0.858 & 0.208 & 0.521 & 0.085 & 0.350 & 0.284 & 0.737 & 0.196 & \worst{0.209} & 0.085 & 0.463 & 0.675 & 0.595 & 0.187 & 0.402 \\
Deepseek VL2-Small & 0.560 & 0.666 & 0.678 & 0.953 & 0.061 & 0.647 & 0.150 & 0.471 & 0.474 & 0.749 & 0.222 & 0.354 & 0.325 & 0.650 & 0.726 & 0.818 & 0.205 & 0.594 \\
Deepseek VL2 & 0.595 & 0.669 & 0.672 & 0.970 & 0.052 & 0.652 & 0.155 & 0.444 & 0.389 & 0.793 & 0.222 & 0.308 & 0.345 & 0.665 & 0.593 & 0.868 & 0.368 & 0.515 \\
InternVL2.5 4B-MPO & 0.415 & 0.637 & 0.638 & 0.898 & 0.178 & 0.573 & 0.145 & 0.405 & 0.375 & 0.788 & 0.175 & 0.296 & 0.140 & 0.566 & 0.871 & 0.633 & 0.115 & 0.551 \\
InternVL2.5 8B-MPO & 0.455 & 0.593 & 0.596 & 0.935 & 0.088 & 0.557 & 0.195 & \best{0.497} & 0.500 & 0.789 & 0.206 & \best{0.395} & 0.440 & 0.740 & 0.662 & 0.897 & 0.433 & 0.594 \\
InternVL3-1B & 0.370 & 0.546 & 0.533 & 0.913 & 0.128 & 0.486 & 0.120 & 0.360 & 0.423 & 0.730 & 0.089 & 0.309 & 0.010 & 0.338 & - & - & - & - \\
InternVL3-2B & 0.420 & 0.645 & 0.665 & 0.891 & 0.156 & 0.593 & 0.135 & 0.380 & 0.419 & 0.753 & 0.111 & 0.316 & 0.140 & 0.534 & 0.255 & 0.815 & 0.437 & 0.208 \\
InternVL3-8B & 0.420 & 0.584 & 0.579 & 0.923 & 0.118 & 0.534 & 0.160 & 0.411 & 0.440 & 0.777 & 0.125 & 0.341 & 0.505 & 0.772 & 0.697 & 0.917 & 0.439 & 0.639 \\
InternVL3-9B & 0.450 & 0.649 & 0.669 & 0.906 & 0.130 & 0.606 & 0.145 & 0.401 & 0.476 & 0.743 & 0.091 & 0.354 & 0.595 & 0.807 & 0.813 & 0.924 & 0.300 & 0.751 \\
InternVL3-14B & 0.525 & 0.652 & 0.654 & 0.947 & 0.097 & 0.619 & 0.150 & 0.436 & 0.439 & 0.764 & 0.180 & 0.335 & 0.585 & 0.820 & 0.805 & 0.921 & 0.402 & 0.742 \\
InternVL3-38B & 0.635 & \best{0.728} & \best{0.733} & 0.965 & 0.075 & \best{0.707} & 0.190 & 0.453 & 0.450 & 0.806 & 0.163 & 0.363 & \best{0.720} & \best{0.875} & 0.821 & \best{0.965} & 0.462 & 0.793 \\
InternVL3.5-1B & 0.335 & 0.516 & 0.517 & 0.897 & 0.109 & 0.464 & 0.075 & 0.352 & 0.482 & 0.628 & 0.096 & 0.303 & 0.150 & 0.497 & 0.562 & 0.718 & 0.215 & 0.404 \\
InternVL3.5-2B & 0.295 & 0.540 & 0.542 & 0.859 & 0.163 & 0.465 & 0.155 & 0.393 & 0.452 & 0.765 & 0.085 & 0.346 & 0.365 & 0.642 & 0.653 & 0.864 & 0.226 & 0.564 \\
InternVL3.5-4B & 0.440 & 0.556 & 0.544 & 0.948 & 0.088 & 0.516 & 0.165 & 0.443 & 0.513 & 0.753 & 0.116 & 0.386 & 0.435 & 0.714 & 0.751 & 0.872 & 0.236 & 0.655 \\
InternVL3.5-8B & 0.395 & 0.554 & 0.531 & 0.928 & 0.129 & 0.493 & 0.135 & 0.398 & 0.440 & 0.744 & 0.125 & 0.328 & 0.495 & 0.752 & 0.721 & 0.908 & 0.349 & 0.655 \\
InternVL3.5-14B & 0.495 & 0.615 & 0.592 & 0.956 & 0.120 & 0.566 & 0.150 & 0.421 & 0.474 & 0.750 & 0.125 & 0.355 & 0.570 & 0.801 & 0.846 & 0.906 & 0.228 & 0.766 \\
InternVL3.5-38B & 0.440 & 0.621 & 0.607 & 0.922 & 0.155 & 0.560 & 0.205 & 0.440 & 0.438 & 0.827 & 0.138 & 0.362 & 0.660 & 0.856 & 0.811 & 0.947 & 0.469 & 0.767 \\
InternVL3.5-1B-Thk & 0.365 & 0.596 & 0.541 & 0.905 & 0.232 & 0.490 & \worst{0.060} & 0.347 & 0.529 & \worst{0.580} & 0.086 & 0.307 & 0.055 & 0.417 & 0.666 & 0.535 & 0.183 & 0.356 \\
InternVL3.5-2B-Thk & 0.255 & 0.586 & 0.571 & 0.815 & 0.282 & 0.465 & 0.065 & 0.345 & 0.463 & 0.612 & 0.115 & 0.283 & 0.225 & 0.591 & 0.515 & 0.805 & 0.364 & 0.414 \\
InternVL3.5-4B-Thk & 0.345 & 0.551 & 0.551 & 0.889 & 0.136 & 0.490 & 0.155 & 0.443 & 0.487 & 0.751 & 0.150 & 0.366 & 0.340 & 0.671 & 0.650 & 0.847 & 0.345 & 0.550 \\
InternVL3.5-8B-Thk & 0.320 & 0.578 & 0.529 & 0.881 & 0.237 & 0.466 & 0.115 & 0.399 & 0.472 & 0.702 & 0.127 & 0.332 & 0.435 & 0.725 & 0.681 & 0.891 & 0.371 & 0.606 \\
InternVL3.5-14B-Thk & 0.440 & 0.635 & 0.624 & 0.916 & 0.168 & 0.572 & 0.145 & 0.420 & 0.490 & 0.738 & 0.115 & 0.361 & 0.520 & 0.789 & 0.842 & 0.886 & 0.270 & 0.746 \\
InternVL3.5-38B-Thk & 0.410 & 0.629 & 0.622 & 0.901 & 0.181 & 0.560 & \best{0.215} & 0.460 & 0.452 & 0.830 & 0.155 & 0.375 & 0.640 & 0.850 & \best{0.874} & 0.924 & 0.332 & \best{0.808} \\
VLM-R1-OVD & \worst{0.200} & \worst{0.509} & 0.544 & \worst{0.778} & 0.188 & 0.423 & 0.120 & 0.420 & 0.535 & 0.688 & 0.111 & 0.368 & 0.270 & 0.621 & 0.620 & 0.810 & 0.314 & 0.502 \\
VLM-R1-Math & 0.335 & 0.556 & 0.560 & 0.879 & 0.145 & 0.492 & 0.145 & 0.426 & 0.518 & 0.727 & 0.103 & 0.376 & 0.320 & 0.649 & 0.706 & 0.820 & 0.238 & 0.579 \\
VLM-R1-REC & 0.390 & 0.611 & 0.641 & 0.883 & 0.125 & 0.566 & 0.105 & 0.404 & \best{0.538} & 0.665 & 0.100 & 0.358 & 0.205 & 0.593 & 0.583 & 0.764 & 0.353 & 0.446 \\
Kimi-VL-A3B-Thk & 0.305 & 0.633 & 0.537 & 0.862 & 0.366 & 0.463 & 0.120 & 0.400 & 0.318 & 0.781 & 0.223 & 0.248 & 0.245 & 0.621 & 0.318 & 0.896 & 0.493  & 0.285 \\
Kimi-VL-A3B-Instruct & 0.370 & 0.545 & 0.540 & 0.910 & 0.117 & 0.491 & 0.120 & 0.379 & 0.355 & 0.762 & 0.168 & 0.270 & 0.425 & 0.682 & 0.703 & 0.881 & 0.211 & 0.620 \\
Kimi-VL-A3B-Thk-2506 & 0.390 & 0.576 & 0.595 & 0.900 & 0.100 & 0.536 & 0.180 & 0.484 & 0.313 & \best{0.863} & 0.311 & 0.270 & 0.450 & 0.719 & 0.640 & 0.912 & 0.375 & 0.584 \\
LLaVA-Interleave-Qwen-0.5B & 0.660 & 0.672 & 0.675 & 0.994 & 0.004 & 0.671 & 0.135 & \worst{0.312} & \worst{0.278} & 0.835 & 0.112 & 0.232 & \worst{0.005} & \worst{0.233} & \worst{0.030} & 0.545 & 0.223 & \worst{0.016} \\
LLaVA-1.5-7B & 0.635 & 0.662 & 0.659 & 0.991 & 0.027 & 0.653 & 0.090 & 0.356 & 0.400 & 0.688 & 0.135 & 0.275 & 0.040 & 0.366 & - & - & - & - \\
LLaVA-Next-Mistral-7B & 0.560 & 0.670 & 0.668 & 0.957 & 0.093 & 0.639 & 0.125 & 0.362 & 0.428 & 0.735 & 0.084 & 0.314 & 0.175 & 0.566 & 0.519 & 0.753 & 0.365 & 0.390 \\
LLaVA-Next-Vicuna-7B & 0.655 & 0.675 & 0.669 & \best{0.995} & 0.028 & 0.666 & 0.145 & 0.372 & 0.415 & 0.769 & 0.091 & 0.319 & 0.080 & 0.482 & 0.095 & 0.825 & 0.447 & 0.078 \\
LLaVA-Interleave-Qwen-7B & 0.490 & 0.647 & 0.615 & 0.945 & 0.173 & 0.581 & 0.155 & 0.394 & 0.437 & 0.772 & 0.100 & 0.337 & 0.020 & 0.360 & 0.719 & \worst{0.403} & 0.251 & 0.290 \\
LLaVA-1.5-13B & 0.495 & 0.620 & 0.605 & 0.951 & 0.112 & 0.576 & 0.120 & 0.393 & 0.461 & 0.714 & 0.118 & 0.329 & 0.090 & 0.458 & 0.563 & 0.630 & 0.235 & 0.355 \\
LLaVA-Next-Vicuna-13B & \best{0.660} & 0.679 & 0.680 & 0.993 & 0.012 & 0.675 & 0.145 & 0.390 & 0.430 & 0.762 & 0.109 & 0.328 & 0.160 & 0.544 & 0.664 & 0.699 & 0.236 & 0.465 \\
Phi-4 & 0.520 & 0.621 & 0.616 & 0.958 & 0.080 & 0.591 & 0.105 & 0.393 & 0.433 & 0.701 & 0.157 & 0.303 & \textcolor{red!70!black}{0.005} & 0.234 & - & - & - & - \\
Qwen2.5VL 3B & 0.265 & 0.516 & \worst{0.459} & 0.871 & 0.216 & \worst{0.400} & 0.130 & 0.415 & 0.494 & 0.716 & 0.121 & 0.354 & 0.250 & 0.603 & 0.627 & 0.793 & 0.282 & 0.497 \\
Qwen2.5VL 7B & 0.465 & 0.621 & 0.623 & 0.929 & 0.111 & 0.579 & 0.165 & 0.424 & 0.472 & 0.769 & 0.115 & 0.363 & 0.580 & 0.819 & 0.835 & 0.912 & 0.348 & 0.761 \\
Qwen2.5VL 32B & 0.375 & 0.589 & 0.617 & 0.883 & 0.115 & 0.545 & 0.205 & 0.459 & 0.487 & 0.805 & 0.130 & 0.392 & 0.650 & 0.851 & 0.865 & 0.930 & 0.344 & 0.805 \\
\bottomrule
\end{tabular}}
\end{table}

\begin{table}[t]
\centering
\caption{Top 12 proprietary and open-source VLMs on Visual Grounding in \pname. 
Only the best value in each group is highlighted; unlisted models show substantially lower RE and $A_{\text{adj}}$.}
\label{tab:detail4}
\setlength{\tabcolsep}{3pt}
\renewcommand{\arraystretch}{0.95}
\scriptsize
\resizebox{\linewidth}{!}{
\begin{tabular}{l|cccccc @{\hskip 10pt} l|cccccc}
\toprule
\multicolumn{7}{c}{\textbf{Proprietary VLMs}} & \multicolumn{7}{c}{\textbf{Open-Source VLMs}} \\
\cmidrule(lr){1-7} \cmidrule(lr){8-14}
\textbf{Model} & \textbf{RE} & \textbf{VE} & $\boldsymbol{\theta}$ & $\boldsymbol{r}$ & $\boldsymbol{g}$ & \textbf{$A_{\text{adj}}$} &
\textbf{Model} & \textbf{RE} & \textbf{VE} & $\boldsymbol{\theta}$ & $\boldsymbol{r}$ & $\boldsymbol{g}$ & \textbf{$A_{\text{adj}}$} \\
\midrule
Gemini 2.5 Pro & 0.280 & 0.716 & \textbf{\textcolor{green!50!black}{0.979}} & 0.731 & 0.012 & 0.716 &
LLaVA-1.5-13B & \textbf{\textcolor{green!50!black}{0.610}} & \textbf{\textcolor{green!50!black}{0.829}} & \textbf{\textcolor{green!50!black}{0.836}} & \textbf{\textcolor{green!50!black}{0.923}} & 0.346 & \textbf{\textcolor{green!50!black}{0.772}} \\
Gemini 1.5 Pro & \textbf{\textcolor{green!50!black}{0.360}} & \textbf{\textcolor{green!50!black}{0.745}} & 0.928 & 0.789 & 0.177 & \textbf{\textcolor{green!50!black}{0.732}} &
LLaVA-1.5-7B & 0.370 & 0.664 & 0.690 & 0.855 & 0.238 & 0.590 \\
Gemini 2.5 Flash & 0.330 & 0.650 & 0.561 & 0.870 & 0.368 & 0.488 &
Qwen2.5VL 32B & 0.405 & 0.589 & 0.580 & 0.914 & 0.140 & 0.530 \\
GPT-4.1 & 0.215 & 0.605 & 0.688 & 0.746 & 0.295 & 0.513 &
LLaVA-Next-Vicuna-13B & 0.285 & 0.590 & 0.571 & 0.839 & 0.259 & 0.479 \\
Gemini 1.5 Flash & 0.220 & 0.450 & 0.483 & 0.822 & 0.103 & 0.397 &
LLaVA-Next-Mistral-7B & 0.305 & 0.573 & 0.446 & 0.906 & 0.304 & 0.404 \\
Gemini 2.5 Flash-Lite & 0.210 & 0.591 & 0.492 & 0.796 & 0.393 & 0.392 &
VLM-R1-OVD & 0.045 & 0.375 & 0.660 & 0.511 & 0.111 & 0.337 \\
GPT-5 & 0.225 & 0.528 & 0.288 & \textbf{\textcolor{green!50!black}{0.926}} & 0.366 & 0.267 &
LLaVA-Next-Vicuna-7B & 0.160 & 0.399 & 0.277 & 0.869 & 0.218 & 0.241 \\
GPT-4.1 Nano & 0.175 & 0.413 & 0.300 & 0.872 & 0.216 & 0.261 &
VLM-R1-Math & 0.035 & 0.330 & 0.533 & 0.506 & 0.130 & 0.269 \\
OpenAI o3 & 0.130 & 0.428 & 0.354 & 0.775 & 0.237 & 0.274 &
VLM-R1-REC & 0.080 & 0.360 & 0.238 & 0.755 & 0.236 & 0.180 \\
Claude 3.5 Sonnet & 0.030 & 0.191 & 0.203 & 0.620 & 0.082 & 0.126 &
Kimi-VL-A3B-Thk-2506 & 0.010 & 0.203 & 0.439 & 0.388 & 0.057 & 0.170 \\
GPT-5 mini & 0.030 & 0.186 & 0.175 & 0.644 & 0.090 & 0.112 &
Deepseek VL2-Small & 0.065 & 0.286 & 0.164 & 0.790 & 0.187 & 0.130 \\
Claude 3.7 Sonnet & 0.035 & 0.136 & 0.144 & 0.702 & 0.041 & 0.101 &
Deepseek VL2-Tiny & 0.030 & 0.161 & 0.095 & 0.750 & 0.100 & 0.071 \\
\bottomrule
\end{tabular}}
\end{table}

\end{document}